\documentclass{article}

\usepackage{microtype}
\usepackage{graphicx}
\usepackage{subcaption}
\usepackage{booktabs} 

\usepackage{hyperref}

\usepackage[preprint]{icml2026}

\usepackage{amsmath}
\usepackage{amssymb}
\usepackage{mathtools}
\usepackage{amsthm}

\usepackage[capitalize,noabbrev]{cleveref}

\theoremstyle{plain}

\theoremstyle{definition}

\theoremstyle{remark}

\usepackage[textsize=tiny]{todonotes}

\usepackage{amsmath,amsfonts,bm}

\def\eqref#1{equation~\ref{#1}}

\def\1{\bm{1}}

\DeclareMathAlphabet{\mathsfit}{\encodingdefault}{\sfdefault}{m}{sl}
\SetMathAlphabet{\mathsfit}{bold}{\encodingdefault}{\sfdefault}{bx}{n}

\usepackage{multirow}
\usepackage{booktabs}
\usepackage{colortbl}
\usepackage{subcaption}

\usepackage{xurl}

\definecolor{lighttable}{RGB}{234, 234, 242}
\definecolor{lightblue}{RGB}{173, 216, 230}
\definecolor{lightcoral}{RGB}{240, 128, 128}
\definecolor{lightgreen}{RGB}{144, 238, 144}
\definecolor{lightskyblue}{RGB}{135, 206, 250}
\definecolor{lightsalmon}{RGB}{255, 160, 122}
\definecolor{palegreen}{RGB}{152, 251, 152}

\definecolor{RoyalBlue}{RGB}{65, 105, 225}
\definecolor{CrimsonRed}{RGB}{220, 20, 60}
\definecolor{DeepPurple}{RGB}{128, 0, 128}
\definecolor{SlateGray2}{RGB}{200, 200, 202}
\definecolor{Olive}{RGB}{153, 128, 0}

\definecolor{Salmon}{RGB}{250, 128, 114}
\definecolor{PaleGreen}{RGB}{152, 251, 152}
\definecolor{LightOrange}{RGB}{255, 204, 153}
\definecolor{PowderBlue}{RGB}{176, 224, 230}
\definecolor{Lavender}{RGB}{230, 230, 250}
\definecolor{LightYellow}{RGB}{255, 255, 204}
\definecolor{Peach}{RGB}{255, 218, 185}
\definecolor{SlateGray}{RGB}{112, 128, 144}
\definecolor{WhiteSmoke}{RGB}{245, 245, 245}

\definecolor{LightCoral}{RGB}{255, 200, 180} %
\definecolor{DarkCoral}{RGB}{205, 90, 80} 
\definecolor{CharcoalGray}{RGB}{50, 50, 50} 

\usepackage{listings}
\usepackage{verbatim}

\usepackage{verbatim}
\newenvironment{itemizeSoheil}
  {\begin{itemize}}
  {\end{itemize}}

\usepackage{ifthen}

\newboolean{comment_summary_bulletpoints}
\setboolean{comment_summary_bulletpoints}{false} 

\newcommand{\blueitemize}[1]{%
  \ifthenelse{\boolean{comment_summary_bulletpoints}}
  {
  } 
  {
    \begingroup
      \renewcommand{\labelitemi}{\textcolor{blue}{\textbullet}}%
      \begin{itemize}
        \color{blue}
        #1
      \end{itemize}
    \endgroup
  } 
}

\newcommand{\soheilhead}[1]{\textbf{#1}:}

\usepackage{xcolor}

\usepackage{dsfont} 
\usepackage{pifont}
\newcommand{\cmark}{\textcolor{green!60!black}{\ding{51}}} 
\newcommand{\xmark}{\textcolor{red}{\ding{55}}}            
\usepackage{ragged2e}   
\usepackage{enumitem} 

\usepackage{adjustbox} 
\usepackage{amssymb}

\usepackage{array}

\usepackage[table,dvipsnames]{xcolor} 
\usepackage{adjustbox}
\usepackage{multirow}

\newcommand{\deltacell}[2]{%
  \begingroup
  \count0=\numexpr#2-#1\relax
  \ifnum\count0>0 \textcolor{green!60!black}{+\the\count0}%
  \else\ifnum\count0<0 \textcolor{red}{\the\count0}%
  \else 0\fi\fi
  \endgroup
}
\usepackage{tikz}
\usetikzlibrary{arrows.meta,positioning,fit,shapes.multipart,backgrounds,calc}
\usepackage{fontawesome5} 

\definecolor{benchBlue}{HTML}{2563EB}         
\definecolor{questionIndigo}{HTML}{4F46E5}    
\definecolor{thesisGreen}{HTML}{16A34A}       
\definecolor{antithesisOrange}{HTML}{F59E0B}  
\definecolor{synthesisTeal}{HTML}{0D9488}     
\definecolor{evalPurple}{HTML}{7C3AED}        
\definecolor{ink}{HTML}{111827}               

\usepackage{fontawesome5}
\usepackage{tikz}
\usetikzlibrary{shapes, arrows.meta, positioning, shadows, fit}
\usetikzlibrary{shadows.blur}

\newcommand{\CardW}{5.5cm}
\newcommand{\CardH}{0.5cm}

\definecolor{thesisGreen}{RGB}{0,128,0}
\definecolor{antithesisOrange}{RGB}{255,140,0}
\definecolor{synthesisTeal}{RGB}{0,128,128}
\definecolor{combinedGreen}{RGB}{0,200,90}
\definecolor{benchBlue}{RGB}{70,130,180}
\definecolor{evalPurple}{RGB}{128,0,128}
\definecolor{axisGray}{gray}{0.3}
\definecolor{ink}{RGB}{30,30,30}

\definecolor{questionBlue}{RGB}{66,139,202}
\definecolor{thesisGreen}{RGB}{40,167,69}
\definecolor{antithesisOrange}{RGB}{230,120,40}
\definecolor{synthesisTeal}{RGB}{0,160,150}
\definecolor{synthesisBlend}{RGB}{153,102,204}
\definecolor{cardBG}{RGB}{245,247,250}
\definecolor{ink}{RGB}{45,45,45}

\tikzset{
  >=Latex,
  every node/.style={font=\small, align=left, text=ink},
  agentA/.style={
    rounded corners=3pt, thick, inner sep=7pt, minimum height=\CardH,
    blur shadow={shadow blur steps=5}, text width=4.5cm,
    draw=thesisGreen!60!black, fill=thesisGreen!12
  },
  agentB/.style={
    rounded corners=3pt, thick, inner sep=7pt, minimum height=\CardH,
    blur shadow={shadow blur steps=5}, text width=4.5cm,
    draw=antithesisOrange!60!black, fill=antithesisOrange!12
  },
  SynthA/.style={
    rounded corners=3pt, thick, inner sep=7pt, minimum height=\CardH,
    blur shadow={shadow blur steps=5}, text width=4.5cm,
    draw=synthesisBlend!60!black, fill=synthesisBlend!12
  },
  papyrusThesis/.style={
    shape=document, thick, inner sep=9pt, minimum height=\CardH,
    draw=thesisGreen!60!black, fill=thesisGreen!7,
    blur shadow={shadow blur steps=5}, text width=4cm
  },
  papyrusAnti/.style={
    shape=document, thick, inner sep=9pt, minimum height=\CardH,
    draw=antithesisOrange!60!black, fill=antithesisOrange!7,
    blur shadow={shadow blur steps=5}, text width=6.6cm
  },
  finalCard/.style={
    rounded corners=3pt, thick, inner sep=7pt, minimum height=\CardH,
    drop shadow={opacity=.14, blur steps=2}, text width=6.9cm,
    draw=synthesisTeal!60!black, fill=synthesisTeal!12
  },
  bench/.style={
    rounded corners=3pt, thick, inner sep=7pt,
    draw=benchBlue!60!black, fill=benchBlue!10, text width=2.5cm,
  },
  evalbox/.style={
    rounded corners=3pt, thick, inner sep=7pt, 
    draw=white!60!black, fill=white!10
  },
  note/.style={
    rounded corners=3pt, semithick, inner sep=6pt, minimum height=\CardH,
    draw=black!30, fill=black!3, text width=5.0cm
  },
  pipe/.style={very thick, -{Latex[length=3mm, width=2mm]}},
  dashedpipe/.style={pipe, dash pattern=on 4pt off 3pt, opacity=.8},
  labelbox/.style={font=\footnotesize, inner sep=2pt, fill=white, text=ink},
}

\tikzset{
  >=Latex,
  font=\small,
  cardBase/.style={
    rectangle, rounded corners=8pt, draw=black!25, fill=cardBG,
    line width=0.8pt, inner sep=8pt, text=ink,
    text width=\CardW, minimum height=\CardH, align=left,
    drop shadow
  },
  qCard/.style={cardBase, draw=black!50!black, fill=questionBlue!4},
  tCard/.style={cardBase, draw=black!50!black, fill=thesisGreen!4},
  aCard/.style={cardBase, draw=black!50!black, fill=antithesisOrange!5},
  sCard/.style={cardBase, draw=black!50!black, fill=synthesisBlend!5},
  pill/.style={
    draw=black!50!black, 
    ,rounded corners=7pt, inner ysep=3.5pt, inner xsep=8pt,
    text=white, font=\bfseries\footnotesize
  },
  qPill/.style={pill, fill=questionBlue},
  tPill/.style={pill, fill=thesisGreen},
  aPill/.style={pill, fill=antithesisOrange},
  sPill/.style={pill, fill=synthesisBlend},
  chip/.style={
    draw=black!50!black, fill=white, rounded corners=5pt, inner xsep=6pt, inner ysep=2.5pt,
    font=\bfseries, text=black
  },
}

\definecolor{thesisGreen}{HTML}{16A34A}        
\definecolor{antithesisOrange}{HTML}{F59E0B}   
\definecolor{synthesisBlend}{RGB}{153,102,204} 
\definecolor{generalBlue}{HTML}{1E3A8A}        

\newcommand{\DialecticFillPct}{8}   
\newcommand{\DialecticFramePct}{65} 
\newcommand{\DialecticTitlePct}{85} 

\colorlet{thesisGreenFill}{thesisGreen!\DialecticFillPct}
\colorlet{thesisGreenFrame}{thesisGreen!\DialecticFramePct!black}
\colorlet{thesisGreenTitle}{thesisGreen!\DialecticTitlePct!black}

\colorlet{antithesisOrangeFill}{antithesisOrange!\DialecticFillPct}
\colorlet{antithesisOrangeFrame}{antithesisOrange!\DialecticFramePct!black}
\colorlet{antithesisOrangeTitle}{antithesisOrange!\DialecticTitlePct!black}

\colorlet{synthesisBlendFill}{synthesisBlend!\DialecticFillPct}
\colorlet{synthesisBlendFrame}{synthesisBlend!\DialecticFramePct!black}
\colorlet{synthesisBlendTitle}{synthesisBlend!\DialecticTitlePct!black}

\colorlet{infoFill}{blue!5}              
\colorlet{infoFrame}{blue!40!black}      
\colorlet{infoTitle}{generalBlue}        

\usepackage{listings}
\lstdefinestyle{promptstyle}{
  basicstyle=\ttfamily\small,
  columns=fullflexible,
  breaklines=true,
  frame=single,
  framerule=0.3pt,
  xleftmargin=1ex,
  xrightmargin=1ex,
  aboveskip=0.6\baselineskip,
  belowskip=0.6\baselineskip,
  tabsize=2
}

\usepackage[most]{tcolorbox}
\tcbset{
  enhanced,
  breakable,
  colback=white,
  colframe=black!10,
  coltitle=black,
  boxsep=1.2ex,
  arc=2pt,
  left=1ex,right=1ex,top=1ex,bottom=1ex
}

\newtcolorbox{infobox}[1]{
  title=\textbf{#1}, breakable,
  colframe=blue!5!black, colback=blue!1,
  colbacktitle=blue!5!black, coltitle=white,
  boxrule=0.8pt, arc=2pt
}

\newtcolorbox{thesisbox}[1]{
  title=\textbf{#1},
  colframe=thesisGreenFrame,  colback=thesisGreenFill,
  colbacktitle=thesisGreenTitle, coltitle=white,
  boxrule=0.8pt, arc=2pt
}
\newtcolorbox{antithesisbox}[1]{
  title=\textbf{#1},
  colframe=antithesisOrangeFrame,  colback=antithesisOrangeFill,
  colbacktitle=antithesisOrangeTitle, coltitle=white,
  boxrule=0.8pt, arc=2pt
}
\newtcolorbox{synthesisbox}[1]{
  title=\textbf{#1},
  colframe=synthesisBlendFrame,  colback=synthesisBlendFill,
  colbacktitle=synthesisBlendTitle, coltitle=white,
  boxrule=0.8pt, arc=2pt
}

\usepackage{tikz}
\usetikzlibrary{arrows.meta,positioning,shapes.geometric,shapes.misc,fit,backgrounds}
\tikzset{
  >={Latex[length=2mm]},
  node/.style={draw, rounded corners, align=center, font=\small, inner sep=6pt},
  process/.style={node, fill=green!6},
  input/.style={node, fill=blue!6},
  branch/.style={diamond, draw, aspect=1.8, align=center, inner sep=3pt, fill=yellow!15},
  agent/.style={node, fill=orange!5},
  output/.style={node, fill=purple!6},
  group/.style={draw=black!30, dashed, rounded corners, inner sep=6pt}
}

\begin{document}

\twocolumn[
  \icmltitle{Evaluating LLM Reasoning Beyond Correctness and CoT}



  \icmlsetsymbol{equal}{*}

  \begin{icmlauthorlist}
    \icmlauthor{Soheil Abbasloo}{comp}
  \end{icmlauthorlist}

  \icmlaffiliation{comp}{Microsoft Research, Vancouver, Canada}

  \icmlcorrespondingauthor{Soheil Abbasloo}{soheil.abbasloo@microsoft.com}

  \icmlkeywords{Machine Learning, ICML}

  \vskip 0.3in
]



\printAffiliationsAndNotice{}  

\begin{abstract}
What does it truly mean for a language model to “reason”? Current evaluations reward models’ correct standalone answers—but correctness alone reveals little about the process that produced them. We argue that reasoning should be understood not as a static chain of steps but as a dynamic trajectory in which ideas interact, clash, and evolve into integrated insights.  
Building on the philosophical tradition of \emph{dialectics}, we introduce \emph{SIEV}, a structured evaluation framework that assesses reasoning through explicit thesis–antithesis–synthesis interactions. SIEV produces interpretable trajectories that highlight key properties of reasoning—robustness to challenge, adaptability under conflict, and synthesis across competing viewpoints—dimensions that conventional correctness-based metrics cannot capture.
Empirical results on GSM and MMLU demonstrate substantial gaps in the reasoning abilities of state-of-the-art models: for example, GPT‑5‑chat loses more than 40 points (out of 100) on GSM when evaluated through SIEV’s process-oriented lens. By shifting focus from \emph{what} answer a model gives to \emph{how} it arrives there, SIEV enables a more transparent and principled distinction between structured reasoning and surface-level pattern generation offering a clearer foundation for assessing and understanding the reasoning capabilities of LLMs.
\end{abstract}

\section{Introduction}
\soheilhead{Reasoning Evaluation Should Evolve}
Reasoning is central to problem-solving and decision-making across scientific analysis, law, education, and collaborative systems, and large language models (LLMs) are increasingly deployed in these settings. This growing reliance suggests that evaluations should capture \emph{how} models reason, not only whether they reach correct outputs. Widely used benchmarks such as GPQA~\cite{gpqa}, MMLU-Pro~\cite{mmlu_pro}, and AIME~\cite{aime2024hf} remain largely outcome-centric, treating reasoning as a static capability inferred from answer correctness. This perspective often obscures the depth, robustness, and internal coherence of the underlying reasoning process, and it can struggle to distinguish genuine reasoning from large-scale pattern imitation acquired during training.

\soheilhead{Limitations of Correctness-Based Evaluation}
Outcome-only metrics provide limited insight into model behavior under uncertainty, ambiguity, or conceptual conflict. Chain-of-thought (CoT) prompting~\cite{wei2022cot} exposes intermediate steps, but CoT traces can be brittle, partially memorized, or insensitive to contradiction. As a result, existing evaluations often cannot clarify whether LLMs perform adaptive, structured inference or rely on surface-level heuristics. This lack of process visibility creates an important interpretability gap: systems may score highly while behaving opaquely, inconsistently, or incoherently in settings that demand reliability. A principled approach to reasoning evaluation should therefore treat reasoning as a \emph{process}, not merely an end product.

\begin{figure*}[t]
    \centering
    \begin{minipage}[b]{0.32\linewidth}
        \centering
         \includegraphics[trim=0.3cm 0.3cm 0.3cm 0.3cm, clip,width=\linewidth]{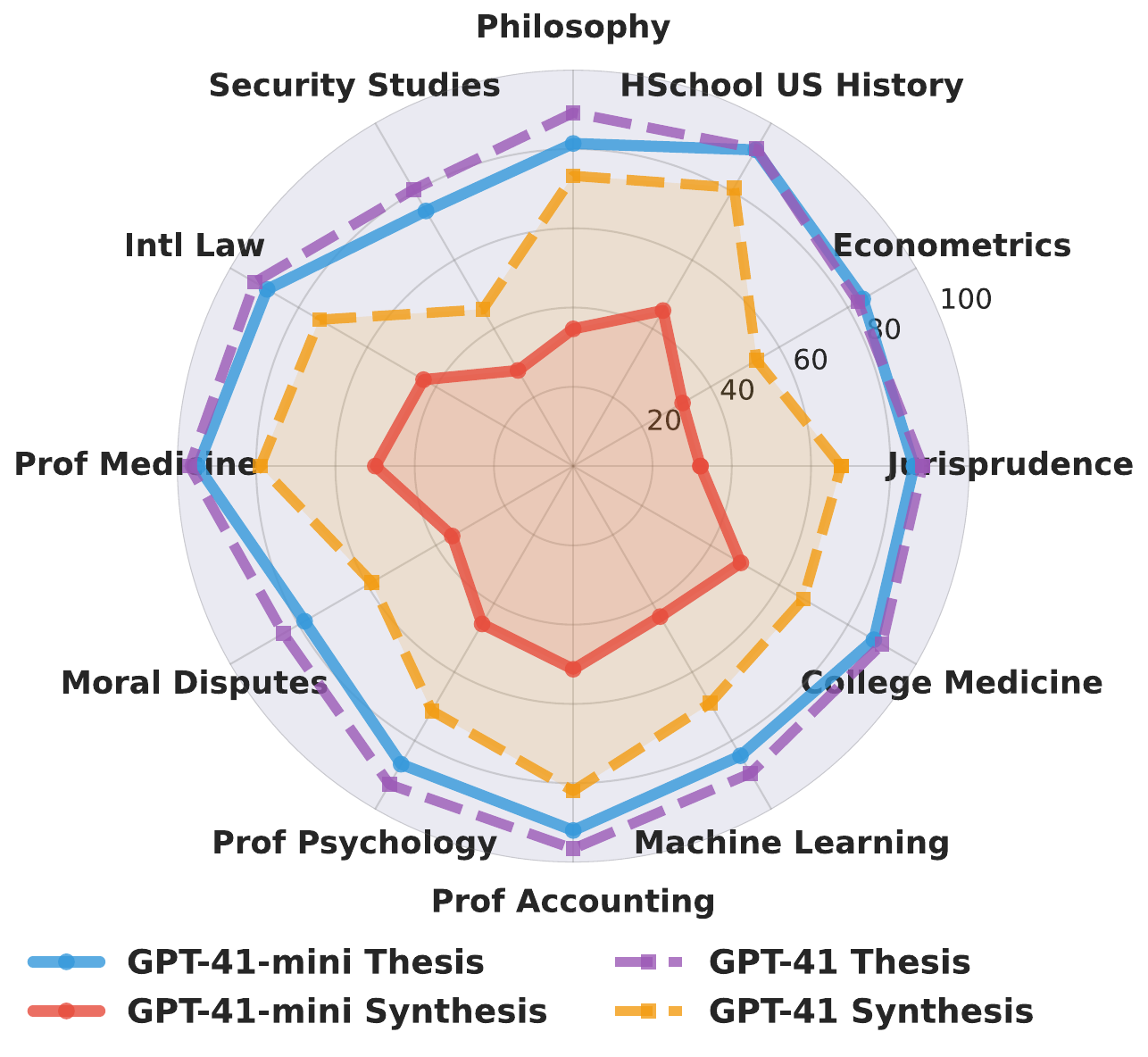}
    \end{minipage}        
    \begin{minipage}[b]{0.32\linewidth}
        \centering
        \includegraphics[trim=0.3cm 0.3cm 0.7cm 0.3cm, clip,width=\linewidth]{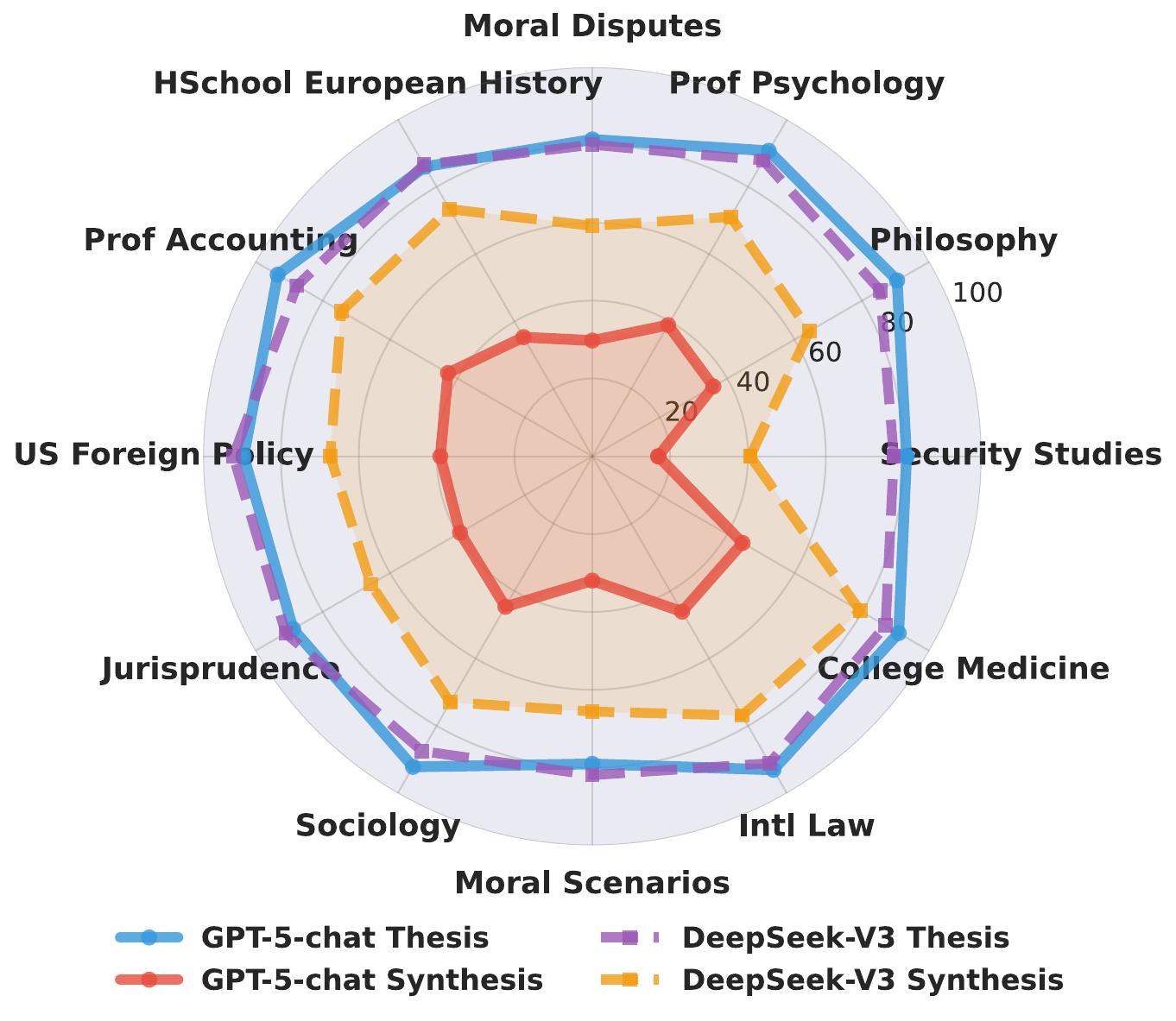}
    \end{minipage}
    \begin{minipage}[b]{0.32\linewidth}
        \centering
        \includegraphics[trim=0.3cm 0.3cm 0.7cm 0.3cm, clip, width=\linewidth]{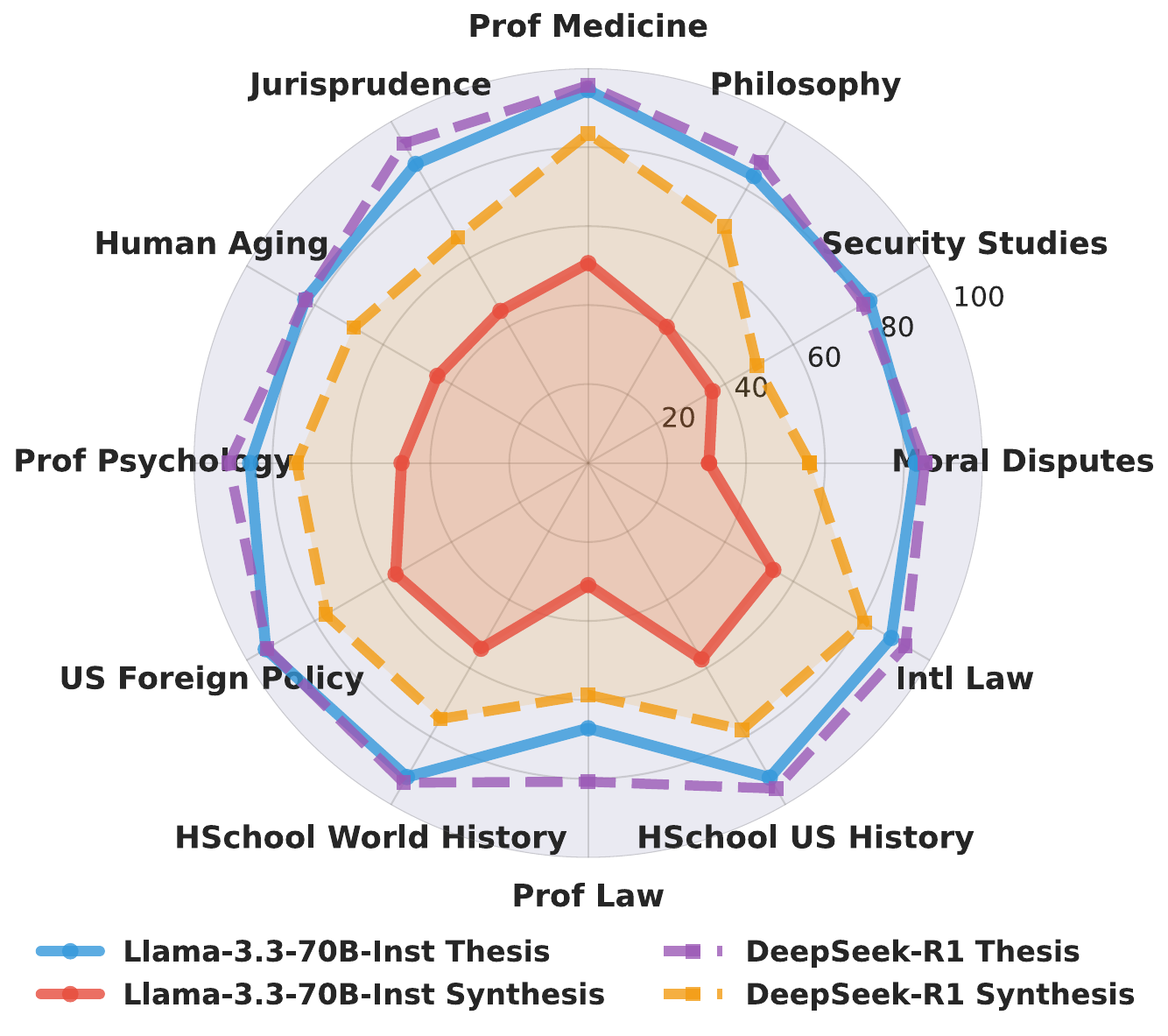}

    \end{minipage}
        \captionsetup{font=normalsize}
        \caption{Sample reasoning performance of models in some MMLU topics. Thesis scores present traditional evaluation of reasoning vs. Synthesis scores indicate our dialectical versions. Our process-driven method reveals significant unreported reasoning gaps between models—differences that standard outcome-oriented evaluations fail to detect and explain.}
    \label{fig:radar:mmlu:sample}
\end{figure*}

\soheilhead{Reasoning as Process}
Human reasoning involves more than producing a correct final answer: it integrates partial evidence, navigates conceptual tension, and reconciles competing viewpoints. Robust reasoning processes can adapt to contradiction while maintaining internal coherence. For LLMs, however, the nature of this adaptability remains unclear: do models plan and strategize, perform causal inference, or largely generate locally plausible sequences? Addressing such questions suggests the need for a structured notion of reasoning that captures iteration, conflict, and synthesis---elements that correctness testing typically leaves implicit.

\soheilhead{Structured Lens for Reasoning}
Among long-studied models of reasoning, \emph{dialectics}—articulated most influentially by Hegel~\cite{hegel_science_logic}—provides precisely such a structure.  
Dialectics frames reasoning as a generative interplay of opposing propositions: a \textit{thesis}, an \textit{antithesis} that challenges it, and a \textit{synthesis} that reconciles the tension between them. 
This triadic pattern foregrounds reasoning as an iterative, tension-resolving process and offers an interpretable lens for evaluating reasoning dynamics beyond surface correctness or sheer CoT fluency.

\soheilhead{The SIEV Framework}
We operationalize this dialectical perspective through SIEV, a structured, process-driven framework for evaluating LLM reasoning. SIEV assesses not only whether a model reaches a correct conclusion, but also whether it: navigates conceptual conflict, sustains competing viewpoints, and synthesizes them into coherent resolutions. SIEV is positioned to reveal weaknesses that correctness-based metrics can miss, including on benchmarks previously viewed as \emph{saturated}.
For example, on MMLU\cite{mmlu}---often treated as broadly solved by top models---SIEV surfaces substantial hidden reasoning gaps (often exceeding 20\%).
Figure~\ref{fig:radar:mmlu:sample} provides an illustrative example: while traditional scores appear competitive, process-oriented scoring highlights cross-model differences in stability, conflict-handling, and synthesis across different MMLU domains.

\soheilhead{Key Contributions of SIEV}
SIEV offers several key advantages:

\noindent\soheilhead{(1) Benchmark- and model-agnostic}  
    SIEV requires no architectural changes, prompt engineering, or benchmark rewriting. It can be applied directly to existing datasets such as GSM8K~\cite{gsm8k} and MMLU, transforming even saturated benchmarks into diagnostics of reasoning behavior.

\noindent\soheilhead{(2) Lower susceptibility to contamination}
    By evaluating reasoning dynamics rather than static answers, SIEV offers lower vulnerability to leakage and template-based CoT artifacts.

\noindent\soheilhead{(3) Exposing hidden weaknesses}  
    SIEV uncovers substantial weaknesses in reasoning robustness and adaptivity that correctness-based metrics and CoT-only analyses overlook (Section~\ref{sec:eval:overall}).
 
\noindent\soheilhead{(4) Natural compatibility with multi-agent systems}  
    As multi-agent LLMs become common, reasoning increasingly emerges from interactions.  
    SIEV naturally extends to evaluating distributed reasoning across multiple models (Section~\ref{sec:eval:cross-model})

\soheilhead{Toward Process-Based Interpretability:}
Grounding evaluation in reasoning processes provides a principled foundation for interpretability. By examining how models integrate evidence, maintain coherence, and resolve conceptual tension, SIEV advances analysis beyond robustness checks or surface-level CoT inspection. This process-oriented perspective helps identify cases where models answer correctly yet fail to engage in stable or adaptive reasoning, supporting more transparent and reliable understanding of model behavior.

\subsection{Related Work: Part II}
\label{appendix:related-work}
The reasoning capabilities of LLMs have been widely studied, with growing skepticism about whether current models genuinely reason or merely imitate reasoning through statistical pattern matching~\cite{dziri2023faith, kambhampati2024can, nezhurina2024alice, mccoy2023embers}. Standard benchmarks such as MMLU~\cite{mmlu} and GSM~\cite{gsm8k} assess correctness of final answers, but this focus often fails to capture the coherence, depth, and adaptability of the underlying reasoning process.

\soheilhead{Heuristic Probing and Its Limits}
Recent work has attempted to expose these limitations by modifying benchmark questions. GSM-Plus~\cite{li2024gsmplus}, GSM-Symbolic~\cite{mirzadeh2024gsmsymbolic}, and functional variants of MATH~\cite{srivastava2024functional} introduce symbolic perturbations or distractors to test sensitivity. Ontology-guided interventions~\cite{hong2024ontology} and causal robustness frameworks~\cite{stolfo2023causal} similarly probe for shallow cue reliance. While these methods reveal fragility, they remain tied to the same paradigm: evaluating correctness under varied surface conditions.

Our work departs from this line by introducing a structured and process-oriented framework grounded in dialectics, SIEV. Drawing on Hegelian philosophy~\cite{hegel_science_logic}, SIEV models reasoning as the interplay of thesis, antithesis, and synthesis. Crucially, SIEV does not alter benchmarks or inject perturbations. Instead, it overlays a formal reasoning scaffold onto existing datasets, making it both \textit{benchmark-agnostic} and \textit{model-agnostic}.

\soheilhead{Reasoning Dynamics over Token Sensitivity}
While prior work has highlighted the fragility of LLM outputs under minimal input changes~\cite{jiang2024peek}, the exponential degradation of multi-step reasoning~\cite{schaeffer2023emergent}, and the correlation between training frequency and test performance~\cite{razeghi2022impact}, these findings point to a broader issue: current models often lack structured reasoning capabilities. Techniques like Chain-of-Thought prompting~\cite{wei2022cot} and scratchpads~\cite{liu2024cot} attempt to organize the reasoning process, but they frequently rely on verbose token generation and still fall short of formal reasoning~\cite{peng2024limitations}.
SIEV offers a complementary perspective. Rather than probing for fragility or relying on token-level cues, it evaluates reasoning as a generative and dialectical process. It focuses on how models navigate conceptual tension and synthesize distinct viewpoints—moving beyond static correctness toward dynamic reasoning evaluation.


\soheilhead{Toward Principled Reasoning Evaluation}
While symbolic and graph-based representations of reasoning~\cite{dziri2023faith} offer valuable insights, they often require task-specific formats or extensive annotation. SIEV, by contrast, provides a general-purpose framework for evaluating reasoning as a dialectical process. It captures not just whether a model arrives at the correct answer, but how it constructs and resolves conceptual tension.

\section{SIEV Framework}
\label{sec:siev-framework}

\subsection{Dialectics as an Evaluation Lens}
We leverage a minimal scaffold from the dialectical tradition~\cite{hegel_science_logic}, framing reasoning as the interplay of a \emph{thesis} (initial stance), \emph{antithesis} (contradictory viewpoint), and \emph{synthesis} (integrated resolution)~\cite{taylorHegel,pinkardHegel,kaufmannExistentialism,beiserHegel}.  
Rather than treating reasoning as a static output, this triadic structure emphasizes iterative, adaptive processes, providing interpretable signals for evaluating LLMs beyond correctness alone or or CoT. Dialectical interaction reveals three complementary aspects of reasoning:

\begin{itemize}[leftmargin=0.5cm,nosep]
    \item \textbf{Coherence:} Maintaining consistent reasoning across steps, avoiding contradictions within the argument.
    \item \textbf{Adaptability:} Revising predictions when faced with meaningful counterpositions, rather than repeating the initial stance.
    \item \textbf{Integration:} Reconciling competing viewpoints into a higher-order synthesis rather than superficial averaging.
\end{itemize}
These signals are invisible to correctness-only metrics and only partially reflected in chain-of-thought traces. Explicitly eliciting thesis–antithesis–synthesis interactions makes reasoning trajectories visible and auditable.

\subsection{The SIEV Pipeline}
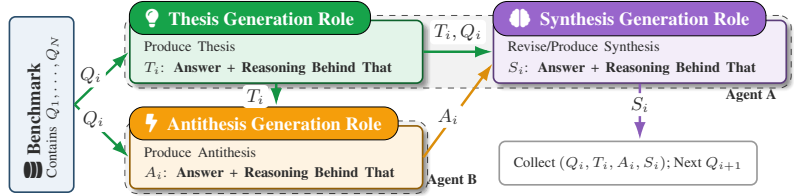
\begin{figure}[t]
  \centering
\resizebox{\columnwidth}{!}{%
\begin{tikzpicture}[x=1cm, y=1cm]

\node[bench, rotate=90] (bench) at (0,0)
{\faDatabase\ \textbf{Benchmark}\\[-2pt]
\scriptsize Contains $Q_1,\dots,Q_N$};

\node[agentA] (agentA) at (4,.9)
{\scriptsize \\Produce Thesis\\$T_i$: \textbf{Answer} + \textbf{Reasoning}\vspace{-3pt}};

\node[agentB] (agentB) at (4,-.9)
{\scriptsize \\Produce Antithesis\\$A_i$: \textbf{Answer} + \textbf{Reasoning}\vspace{-3pt}};

\node[SynthA] (agentSynth) at (10.2,.9)
{\scriptsize \\Revise/Produce Synthesis\\$S_i$: \textbf{Answer} + \textbf{Reasoning}\vspace{-3pt}};

\begin{scope}[on background layer]
  \node[
    draw=gray!60!black, dashed, rounded corners=6pt, inner sep=2pt,
    fill=gray!10,
    fit=(agentA)(agentSynth),
    label={[font=\scriptsize, anchor=south, xshift=5cm, yshift=-15pt]south :\textbf{Agent A}}
  ] (sharedLLM) {};
\end{scope}

\begin{scope}[on background layer]
  \node[
    draw=gray!60!black, dashed, rounded corners=6pt, inner sep=2pt,
    fill=gray!10,
    fit=(agentB),
    label={[font=\scriptsize, anchor=south, xshift=3.3cm, yshift=-0.1pt]south :\textbf{Agent B}}
  ] (sharedLLMB) {};
\end{scope}

\node[tPill, anchor=north west] at ($(agentA.north west)+(0pt,9pt)$) {\faLightbulb\; Thesis Generation Role};
\node[aPill, anchor=north west] at ($(agentB.north west)+(0pt,9pt)$) {\faBolt\; Antithesis Generation Role};
\node[sPill, anchor=north west] at ($(agentSynth.north west)+(0pt,9pt)$) {\faBrain\; Synthesis Generation Role};

\node[evalbox] (eval) at (10.5,-1)
{\scriptsize Collect $(Q_i, T_i, A_i, S_i)$; Next $Q_{i+1}$};

\draw[pipe, thesisGreen!90!black] (bench.south) --  node[pos=0.3, yshift=0pt, xshift=0pt, above, labelbox,fill=white, rounded corners=2pt]{$Q_i$} (agentA.west);
\draw[pipe, thesisGreen!90!black] (bench.south) --  node[pos=0.35, yshift=-5pt, xshift=0pt, above, labelbox ,fill=white, rounded corners=2pt]{$Q_i$} (agentB.west);
\draw[pipe, thesisGreen!90!black] (agentA) -- node[pos=0.5, yshift=-7pt, xshift=-10pt, above, labelbox ,fill=white, rounded corners=2pt]{$T_i$} ($(agentB)+(0pt,25pt)$);
\draw[pipe, thesisGreen!90!black] (agentA) -- node[pos=0.5, yshift=3pt, above, labelbox ,fill=white, rounded corners=2pt]{$T_i, Q_i$} (agentSynth);
\draw[pipe, antithesisOrange!90!black] (agentB.east) -- node[pos=0.3, xshift=2pt, above, labelbox ,fill=white, rounded corners=2pt]{$A_i$} ($(agentSynth.west)-(0,0.2)$);
\draw[pipe, synthesisBlend!90!black] (agentSynth.south) -- node[pos=0.6, above, labelbox ,fill=white, rounded corners=2pt]{$S_i$} (10.2,-0.6); 

\end{tikzpicture}
}
    \caption{
    Overview of SIEV and its TAS pipeline. 
    }   
  \label{fig:sieve:pipeline}
\end{figure}

We introduce SIEV, a practical framework for operationalizing dialectical evaluation.  
SIEV enforces explicit reasoning steps and produces interpretable artifacts that stress-test reasoning under conceptual tension. The pipeline, illustrated in Figure~\ref{fig:sieve:pipeline}, proceeds in three stages.

\textbf{Stage 1 - Thesis:} The model answers a question $Q_i$ to produce a thesis $T_i$, comprising (1) an answer and (2) supporting reasoning.

\textbf{Stage 2 - Antithesis:} The model generates a contradictory response $A_i$, including its reasoning, introducing structured conceptual tension.

\textbf{Stage 3 - Synthesis:} Finally, the model reconciles $T_i$ and $A_i$ into a synthesis $S_i$, producing a refined conclusion that integrates both perspectives.

\textbf{Implementation:} A practical implementation can employ two agents instantiated from the same model: one responsible for thesis and synthesis, and the other for antithesis. This preserves independence of the opposing viewpoints while keeping the approach model-agnostic. Without loss of generality, we adopt this two-agent terminology throughout the paper. Appendix~\ref{app:prompts} provides more details on the prompts used. 

\newcommand{\Problem}{
A cup of mushrooms weighs 100 grams and has 3 grams of protein. If John eats 200 grams of mushrooms every day, how many grams of protein does he get per week?
}

\newcommand{\Answer}{
\textbf{Answer: 42}
}

\newcommand{\ThesisOne}{
\textbf{Reasoning:}\\
A cup of mushrooms (\(100\,\text{g}\)) contains \(3\,\text{g}\) of protein.  
John eats \(200\,\text{g}\) of mushrooms each day, which is twice \(100\,\text{g}\), so he gets:
$2 \times 3\,\text{g} = 6\,\text{g of protein per day}$. Over a week (\(7\) days), the total protein intake is:$ 7 \times 6\,\text{g} = 42\,\text{g}$ $\quad$ \textbf{Answer: 42}
}

\newcommand{\ThesisTwo}{
\textbf{Reasoning:}\\
First, determine the protein per gram of mushrooms. Since 100 grams have 3 grams of protein, each gram provides:$\frac{3}{100} = 0.03$ grams of protein. John eats 200 grams daily, so daily protein intake is:
$200 \times 0.03 = 6$ grams. Over a week (7 days), total protein is: $6 \times 7 = 42$ grams.
\quad \textbf{Answer: 42}
}

\newcommand{\AntiThesisOne}{
\textbf{Reasoning:}
The provided solution is a textbook exercise in oversimplification.
\begin{enumerate}[label=\arabic*., leftmargin=*, nosep]
  \item It accepts the ``3 g per 100 g'' label as exact. USDA data list \(\,3.1\,\text{g}\) protein per \(100\,\text{g}\) raw white mushroom; the \(3\,\text{g}\) figure is already rounded, so linearly scaling it compounds rounding error.
  \item It supposes protein scales perfectly with weight, ignoring that mushrooms lose \(\approx 15\%\) water when saut\'e{}ed or grilled---John is unlikely to eat them raw every time. Cooking concentrates protein, so multiplying raw weights by a raw-based ratio misstates the true intake.
  \item It blindly multiplies by seven, assuming John never skips a day. In real nutrition tracking a ``week'' commonly records six identical intake days plus one free or varied day. Without explicit wording, treating ``every day'' as ``seven identical days'' is an unverified leap.
  \item Volume (a cup) and mass (grams) are conflated. A cup is a measure of volume, not weight, and actual cup weights vary by mushroom type; using a single figure (\(100\,\text{g}\)) as if it were universal ignores natural variability.
\end{enumerate}
A more defensible calculation is:
  1) adopt the more accurate \(3.1\,\text{g}/100\,\text{g}\) raw density, 2) adjust for average \(15\%\) moisture loss in cooking (\(200\,\text{g}\) raw \(\to 170\,\text{g}\) cooked), 3) assume six repeat intake days per common dietary week.
$\text{Daily protein} \;\approx\; \frac{3.1~\text{g}}{100~\text{g}} \times 170~\text{g} \;\approx\; 5.27~\text{g}.$
$\text{Weekly protein} \;\approx\; 6 \times 5.27~\text{g} \;\approx\; 31.6~\text{g}, \text{ rounded to } 32~\text{g}.$

Thus the claimed \(42\,\text{g}\) is overstated by roughly a third. $\quad$ \textbf{Answer: 32}
}

\newcommand{\AntiThesisTwo}{
\textbf{Reasoning:}\\
The original answer assumes unwarranted precision. The protein value (3g) has only one significant figure, while the quantities 100g and 200g are treated as exact. Strict adherence to significant figures requires rounding the final result to match the least precise data point. Thus, although:
$6\ \text{g/day} \times 7\ \text{days} = 42\ \text{g/week}$
the value 42 should be rounded to one significant figure, yielding: $\boxed{40\ \text{grams per week}}$
This reframes the answer to emphasize scientific rigor over arithmetic exactness.
\textbf{Answer: 40}
}

\newcommand{\SynthesisOne}{
\textbf{Reasoning:}\\
The problem is purely mathematical and explicitly states:
\begin{itemize}[leftmargin=*,nosep]
  \item \(100\,\text{g}\) mushrooms \(\to 3\,\text{g}\) protein (this is to be treated as an exact ratio for the purpose of the exercise).
  \item John eats \(200\,\text{g}\) of mushrooms every day.
  \item ``Every day'' means each of the seven days in a week.
\end{itemize}
Given those stated facts:
  (1) Protein per gram \(= \frac{3\,\text{g}}{100\,\text{g}} = 0.03\,\text{g/g}\).
  (2) Daily intake \(= 200\,\text{g} \times 0.03\,\text{g/g} = 6\,\text{g protein}\).
  (3) Weekly intake \(= 6\,\text{g/day} \times 7\,\text{days} = 42\,\text{g protein}\).

The reviewer’s objections (rounding differences, cooking water loss, possible rest days, cup-to-gram variability) introduce assumptions that are not part of the original statement and therefore fall outside the scope of the calculation requested. Staying strictly within the given information yields \(42\,\text{g}\). \quad \textbf{Answer: 42}
}

\newcommand{\SynthesisTwo}{
\textbf{\textless think\textgreater} Okay, let's tackle this again. The original answer calculated 42 grams of protein per week. But Reviewer 1 pointed out that the problem involves significant figures. Let me check that. The problem states: 3 grams of protein (1 significant figure) and 100 grams (which could be 1, 2, or 3 significant figures depending on context, but since it's "100" without a decimal, it's ambiguous). However, in many cases, trailing zeros without a decimal are considered not significant. So 100 grams might be 1 significant figure. But 3 grams is definitely 1 significant figure. If we follow significant figure rules strictly, the least precise measurement (1 significant figure) dictates the final answer's precision. So 200 grams (could be 2 or 3 significant figures, but likely treated as exact here) times 3 grams per 100 grams gives 6 grams per day (1 significant figure). Then: $6\ \text{grams/day} \times 7\ \text{days} = 42\ \text{grams/week}$ But since 6 has 1 significant figure, the answer should be rounded to: $\boxed{40\ \text{grams/week}}$ However, another perspective is that 100 grams and 200 grams are exact quantities (e.g., defined values), so their significant figures don't limit precision. But the 3 grams is a measured value with 1 significant figure. In that case, all calculations should follow the 1 significant figure rule. So: $\frac{3}{100} = 0.03\ \text{g/g} \quad (1\ \text{sig fig}), \quad 200 \times 0.03 = 6\ \text{g/day} \quad (1\ \text{sig fig}), \quad 6 \times 7 = 42 \Rightarrow \boxed{40\ \text{g/week}}$ Thus, the correct answer considering significant figures is: $\boxed{40\ \text{grams}}$\textbf{\textless /think\textgreater}\\
\textbf{Reasoning:}\\
The protein value (3g) has 1 significant figure, which limits the precision of the result. Calculations yield 6g daily (1 sig fig), leading to:
$6 \times 7 = 42 $ Rounded to 1 significant figure, the final answer is: $\boxed{40\ \text{grams}}$\quad \textbf{Answer: 40}
}

\begin{figure*}[t]
\centering
\resizebox{\textwidth}{!}{%
\begin{tikzpicture}[x=1cm, y=1cm]

\node[qCard, text width=24cm] (Q) at (-1,0) {
  \footnotesize
  \justify
  \Problem
};

\node[tCard, below=1.2cm of Q.south west, anchor=west, text width=12.3cm] (T) {
  \footnotesize
  \justify  
  \ThesisOne
};

\node[tCard, below=1.4cm of Q.south east, anchor=east, text width=10.7cm] (T2) {
  \footnotesize
  \justify
  \ThesisTwo
};

\node[aCard, below=3.5cm of T.south west, anchor=west, text width=12.3cm] (A) {
  \footnotesize
  \justify
  \AntiThesisOne
};

\node[aCard, below=2cm of T2.south west, anchor=west, text width=10.7cm] (A2) {
  \footnotesize
  \justify  
  \AntiThesisTwo
};

\node[sCard, below=6cm of A.east, anchor=east, text width=12.3cm] (S) {
  \footnotesize
  \justify
  \SynthesisOne
};

\node[sCard, below=6.7cm of A2.west, anchor=west, text width=10.7cm] (S2) {
  \footnotesize
  \justify
  \SynthesisTwo
};

\node[coordinate, below=.3cm of S.south] (Sbottom) {};
\node[coordinate, below=.3cm of S2.south] (Sbottom2) {};

\begin{pgfonlayer}{background}
  \node[draw=black!40, fill=gray!5, dashed, rounded corners, fit=(T)(A)(S)(Sbottom), inner sep=0.15cm] (back1) {};
  \node[draw=black!40, fill=gray!5, dashed, rounded corners, fit=(T2)(A2)(S2)(Sbottom2), inner sep=0.15cm] (back2) {};
\end{pgfonlayer}

\node at ($(Sbottom.south) + (0,+2pt)$) {\textbf{O3 Model}};
\node at ($(Sbottom2.south) + (0,+2pt)$) {\textbf{DeepSeek-R1 Model}};

\node[qPill, anchor=north west] at ($(Q.north west)+(0pt,7pt)$) {\faQuestionCircle\; Question};
\node[tPill, anchor=north west] at ($(T.north west)+(0pt,9pt)$) {\faLightbulb\; Thesis};
\node[aPill, anchor=north west] at ($(A.north west)+(0pt,9pt)$) {\faBolt\; Antithesis};
\node[sPill, anchor=north west] at ($(S.north west)+(0pt,9pt)$) {\faBrain\; Synthesis};
\node[tPill, anchor=north west] at ($(T2.north west)+(0pt,9pt)$) {\faLightbulb\; Thesis};
\node[aPill, anchor=north west] at ($(A2.north west)+(0pt,9pt)$) {\faBolt\; Antithesis};
\node[sPill, anchor=north west] at ($(S2.north west)+(0pt,9pt)$) {\faBrain\; Synthesis};

\node[chip, draw=black!50!black, anchor=north east] at ($(Q.south east)+(-6pt,35pt)$) {\Answer};
\node[chip, draw=black!50!black, anchor=north east] at ($(T.south east)+(-6pt,16pt)$) {\cmark};
\node[chip, draw=black!50!black, anchor=north east] at ($(A.south east)+(-6pt,16pt)$) {\xmark};
\node[chip, draw=black!50!black, anchor=north east] at ($(S.south east)+(-6pt,16pt)$) {\cmark};
\node[chip, draw=black!50!black, anchor=north east] at ($(T2.south east)+(-6pt,16pt)$) {\cmark};
\node[chip, draw=black!50!black, anchor=north east] at ($(A2.south east)+(-6pt,16pt)$) {\xmark};
\node[chip, draw=black!50!black, anchor=north east] at ($(S2.south east)+(-6pt,16pt)$) {\xmark};

\node[font=\small, text=black, align=left, below=1.2cm of S.south west, anchor=west, text width=13cm, xshift=-.2cm] {
    \justify 
    {
    \textbf{Notes}: For brevity, here, we remove the thinking tokens of R1 in thesis/antithesis. In SIEV, we redact thinking tokens when passing them for different reasons: (1) This helps to keep context length comparable to other models and have fair comparisons, (2) the thinking block of antithesis often reveals correct answer even when it generates an opposing one later!
    }
};

\end{tikzpicture}%
}
\caption{An illustration of dialectical reasoning evaluation in SIEV using two sample LLMs}
\label{fig:siev:example:o3-r1}
\end{figure*}

\subsection{Illustrative Example}
Figure~\ref{fig:siev:example:o3-r1} shows example traces on a simple nutrition-based arithmetic question from GSM~\cite{cobbe2021training}.  
Both models initially produce similar theses, CoTs, and correct final answers. So, a conventional evaluation would treat them equally and concludes that both models gain a similar reasoning quality with respect to this question. However, SIEV continues. 
Model O3 generates a rich antithesis that challenges assumptions behind the original calculation and then synthesizes both views, reaffirming the original answer with justification.
In contrast, R1 produces a more modest antithesis and even with this modest version, after a lengthy verbose sequence of tokens, it fails to reconcile the tension, defaulting to a shallow restatement of the antithesis without deeper reasoning and justification. 

This trace illustrates how SIEV can provide a deeper picture for explaining models' outputs, and differentiate models based on robustness and depth of reasoning, beyond correctness-only. By enforcing structured interaction, SIEV can reveal the depth and robustness of models reasoning and whether it collapses into reasoning pattern mimicry. Appendix~\ref{app:siev:examples} presents another illustrative example using GPT-5 model.

\rowcolors{4}{lighttable!100}{white}
\begin{table*}[t]
\centering
\caption{Overall dialectical results with ranking based on different metrics. Notes: (1) Numbers are in percentage points. (2) DS (\%) computed as \(p_S \times (\lambda + (1-\lambda)) \cdot \text{OC}^\gamma)\) with {\(\gamma=1\)} and {\(\lambda=0.7\)}. (3) Rankings use dense ranking (1, 1, 2, 3). (4) Ranking ties are assigned when absolute score differences are $\le 0.5$. (5) The top 3 ranks with respect to $p_T$, $p_S$, and DS are made bold.}
\scriptsize
\setlength{\tabcolsep}{4pt}
\begin{adjustbox}{max width=\textwidth}
\begin{tabular}{|l|lllll|llllll|}
\hline
\multirow{2}{*}{\textbf{Model}} & \multicolumn{5}{c|}{\textbf{GSM}} & \multicolumn{6}{c|}{\textbf{MMLU}} \\
\cline{2-12}
& $p_T$ (Rank) & $p_S$ (Rank) & $\Delta$ & OC & DS (Rank) &
& $p_T$ (Rank) & $p_S$ (Rank) & $\Delta$ & OC & DS (Rank) \\
\hline
O3          & \textbf{97.1$\pm$0.1 (1)} & \textbf{93.6$\pm$0.7 (1)} & -3.5 & 95.5 & \textbf{92.3 (1)} &
& \textbf{92.2$\pm$0.1 (1)} & \textbf{90.3$\pm$0.2 (1)} & -1.9 & 92.7 & \textbf{88.4 (2)} \\
GPT-4       & {95.1$\pm$0.2 (4)} & \textbf{88.3$\pm$0.5 (3)} & -6.8 & 83.1 & \textbf{84.3 (2)} &
& {86.0$\pm$1.3 (7)} & {75.8$\pm$1.3 (7)} & -10.2 & 93.7 & 74.4 (7) \\
O1          & \textbf{96.5$\pm$0.2 (2)} & 82.7$\pm$1.0 (5) & -13.8 & 97.0 & \textbf{82.0 (3)} &
& \textbf{91.1$\pm$0.0 (2)} & \textbf{90.3$\pm$0.1 (1)} & -0.7 & 96.0 & \textbf{89.2 (1)} \\
Kimi-K2     & \textbf{96.8$\pm$0.2 (1)} & 81.6$\pm$1.1 (6) & -15.2 & 92.5 & {79.7 (4)} &
& {88.6$\pm$0.1 (5)} & 80.7$\pm$0.2 (4) & -7.9 & 94.0 & {79.3 (4)} \\
GPT-5       & \textbf{96.8$\pm$0.2 (1)} & 82.0$\pm$0.8 (6) & -14.7 & 81.4 & {77.5 (5)} &
& \textbf{92.2$\pm$0.1 (1)} & \textbf{86.7$\pm$0.1 (2)} & -5.5 & 92.5 & \textbf{84.8 (3)} \\
Llama-3.3-70B-Inst
            & \textbf{96.2$\pm$0.4 (2)} & \textbf{89.5$\pm$0.1 (3)} & -6.7 & 51.9 & 76.6 (6) &
& 85.1$\pm$0.1 (6) & 58.3$\pm$0.1 (14) & -26.8 & 80.7 & 54.9 (14) \\
DeepSeek-R1 & \textbf{96.1$\pm$0.2 (2)} & 79.2$\pm$1.0 (8) & -16.8 & 75.8 & 73.5 (6) &
& \textbf{90.1$\pm$0.1 (3)} & 77.3$\pm$0.4 (6) & -12.8 & 75.1 & 71.5 (9) \\
GPT-5-mini  & \textbf{96.9$\pm$0.1 (1)} & 80.1$\pm$0.5 (7) & -16.8 & 71.4 & 73.3 (6) &
& {89.3$\pm$0.1 (4)} & 79.0$\pm$0.3 (5) & -10.3 & 84.3 & 75.3 (6) \\
O3-mini     & \textbf{96.1$\pm$0.3 (2)} & \textbf{91.6$\pm$0.3 (2)} & -4.5 & 29.4 & 72.2 (7) &
& 85.0$\pm$0.1 (8) & \textbf{83.8$\pm$0.1 (3)} & -1.3 & 55.0 & 72.4 (8) \\
GPT-4o      & \textbf{95.9$\pm$0.3 (3)} & {84.2$\pm$0.3 (4)} & -11.7 & 48.4 & 71.2 (8) &
& {86.5$\pm$0.1 (6)} & 68.6$\pm$0.2 (11) & -17.9 & 93.7 & 67.3 (12) \\
DeepSeek-V3 & \textbf{96.3$\pm$0.2 (2)} & 75.3$\pm$0.3 (9) & -21.0 & 69.8 & 68.4 (9) &
& {86.6$\pm$0.2 (6)} & {73.8$\pm$0.2 (8)} & -12.8 & 86.7 & 70.8 (10) \\
O4-mini     & \textbf{96.5$\pm$0.2 (2)} & 69.4$\pm$0.6 (11) & -27.1 & 94.9 & 68.4 (9) &
& {88.6$\pm$0.1 (5)} & 77.7$\pm$0.2 (6) & -10.9 & 93.4 & {76.1 (5)} \\
GPT-4.1      & {95.1$\pm$0.4 (4)} & 70.3$\pm$0.2 (10) & -24.8 & 90.1 & 68.2 (9) &
& {88.7$\pm$0.0 (5)} & 73.2$\pm$0.2 (9) & -15.6 & 95.3 & 72.1 (8) \\
Phi-4       & {94.2$\pm$0.3 (5)} & {84.5$\pm$0.8 (4)} & -9.7 & 35.2 & 68.1 (9) &
& 82.0$\pm$0.1 (9) & {73.5$\pm$0.2 (8)} & -8.5 & 81.1 & 69.3 (11) \\
GPT-4.1-mini & \textbf{95.5$\pm$0.0 (3)} & 70.4$\pm$1.1 (10) & -25.1 & 77.8 & 65.7 (10) &
& 84.9$\pm$0.1 (8) & 54.7$\pm$0.2 (13) & -30.2 & 91.0 & 53.2 (15) \\
GPT-5-nano  & {94.8$\pm$0.4 (4)} & {84.4$\pm$0.8 (4)} & -10.4 & 15.5 & 63.0 (11) &
& 85.8$\pm$0.2 (7) & 62.4$\pm$0.1 (13) & -23.4 & 43.3 & 51.8 (16) \\
O1-mini     & 93.3$\pm$0.4 (6) & 59.6$\pm$1.1 (13) & -33.7 & 78.7 & 55.8 (12) &
& 81.8$\pm$0.2 (9) & 69.9$\pm$0.1 (10) & -11.9 & 89.0 & 67.6 (12) \\
Qwen2.5-72B-Inst
            & \textbf{95.7$\pm$0.2 (3)} & 63.8$\pm$0.3 (12) & -31.9 & 56.7 & 55.5 (12) &
& 84.4$\pm$0.1 (8) & 64.8$\pm$0.2 (12) & -19.7 & 64.7 & 57.9 (13) \\
GPT-5-chat  & \textbf{96.4$\pm$0.1 (2)} & 56.2$\pm$1.0 (14) & -40.2 & 85.2 & 53.7 (13) &
& {88.2$\pm$1.7 (5)} & 50.5$\pm$0.9 (15) & -37.7 & 95.1 & 49.8 (17) \\
Ministral-8B-Inst-2410
            & 86.1$\pm$0.9 (7) & 55.9$\pm$0.3 (14) & -30.2 & 63.8 & 49.8 (14) &
& 62.3$\pm$0.2 (11) & 41.3$\pm$0.3 (16) & -20.9 & 59.3 & 36.3 (18) \\
GPT-3.5     & 76.4$\pm$0.6 (7) & 40.2$\pm$0.6 (15) & -36.2 & 50.0 & 34.5 (15) &
& 67.7$\pm$0.8 (10) & 35.2$\pm$0.3 (17) & -32.4 & 65.4 & 31.6 (19) \\
\hline
\end{tabular}
\end{adjustbox}
\label{table:eval:overall}
\end{table*}

\subsection{Evaluation Metrics}
We define three complementary metrics to quantify dialectical reasoning:

  \noindent\textbf{Synthesis Score} ($p_S$): Accuracy after reconciliation, reflecting reasoning quality post-dialectical process.
  
\noindent\textbf{Dialectic Score} (DS): Combines synthesis quality with opposition strength:
    {\footnotesize
    \[
    DS = p_S \times (\lambda + (1-\lambda) p_{\mathrm{OC}}^{\,\gamma})
    \]
    }
    where $p_{\mathrm{OC}}$ denotes opposition compliance (OC: fraction of items where thesis and antithesis differ), $\lambda \in [0,1]$ balances synthesis vs opposition, and $\gamma \ge 0$ shapes curvature of the opposition bonus. DS rewards models that generate meaningful antitheses and robust synthesis.
  
  \noindent\textbf{Improvement} ($\Delta$): Gain from thesis to synthesis:
    $    \Delta = \mathbb{E}[p_S - p_T]$
    where $p_T$ is the conventional thesis score. Positive $\Delta$ signals reasoning evolved beyond initial stance—evidence of adaptability.

Together, these metrics shift evaluation from static correctness toward dynamic reasoning capability, offering a principled bridge between token-level prediction and structured reasoning. 
Although these signals do not necessarily certify authentic reasoning, they offer \emph{stronger grounds} for interpreting whether and how a model’s apparent reasoning reflects a \emph{stable, integrative process} or a \emph{fragile, replay-driven trajectory}.

\section{Evaluation}
\label{sec:eval}

\soheilhead{Setting} 
We evaluate on two saturated benchmarks, GSM8K~\cite{gsm8k} and {MMLU}~\cite{mmlu}, spanning topics from U.S. foreign policy to high‑school science and mathematics. Both benchmarks yield near‑ceiling accuracy for several state‑of‑the‑art models, making it difficult to differentiate reasoning capabilities via static metrics alone. SIEV adds {process‑level signals} (coherence across steps, opposition engagement, and adaptability under contradiction) that can help explain reasoning quality with more transparency. Our study covers {21 LLMs}, ranging from small to large, proprietary to open‑source (the SIEV source code is publicly available at \url{https://github.com/microsoft/siev}.).

\subsection{Overall Results}
\label{sec:eval:overall}
Table~\ref{table:eval:overall} reports SIEV signals across the 21 evaluated models on GSM8K and MMLU. As expected from saturated benchmarks, \emph{thesis} accuracy ($p_T$) clusters near ceiling for many systems (e.g., O3, GPT‑5‑chat, Kimi‑K2 each exceed 96 on GSM8K). In contrast, the \emph{process-level signals} diverge substantially: $p_S$ spans a much wider range, from above 90 (O3, O1) to below 60 (GPT‑5‑chat, Ministral‑8B‑Inst), OC varies from consistently high (O1, O3) to very low (GPT‑5‑nano, O3‑mini), etc.
These differences indicate that similar static correctness can mask very different trajectories—some \emph{stable and integrative}, others \emph{fragile or replay-like}.

\begin{figure*}[!tp]
    \centering
    \begin{minipage}[b]{0.45\textwidth}
        \centering
        \includegraphics[trim=0cm 0cm 0cm 0cm, clip,width=\textwidth]{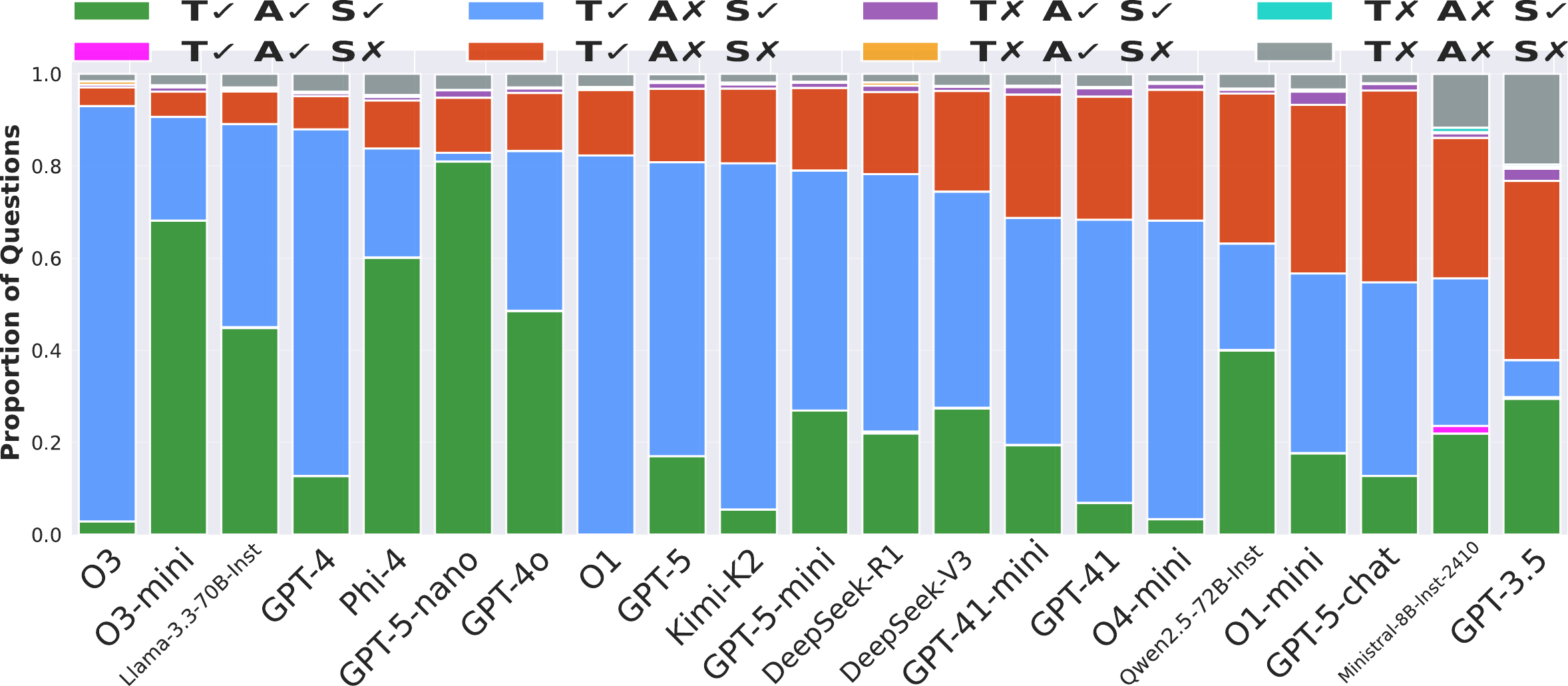}
        \captionsetup{font=tiny} 
        \caption*{
          \parbox{\textwidth}{
            \centering
            \textbf{GSM}
            }
        }
    \end{minipage}   
    \begin{minipage}[b]{0.45\textwidth}
        \centering
        \includegraphics[trim=0cm 0cm 0cm 0cm, clip,width=\textwidth]{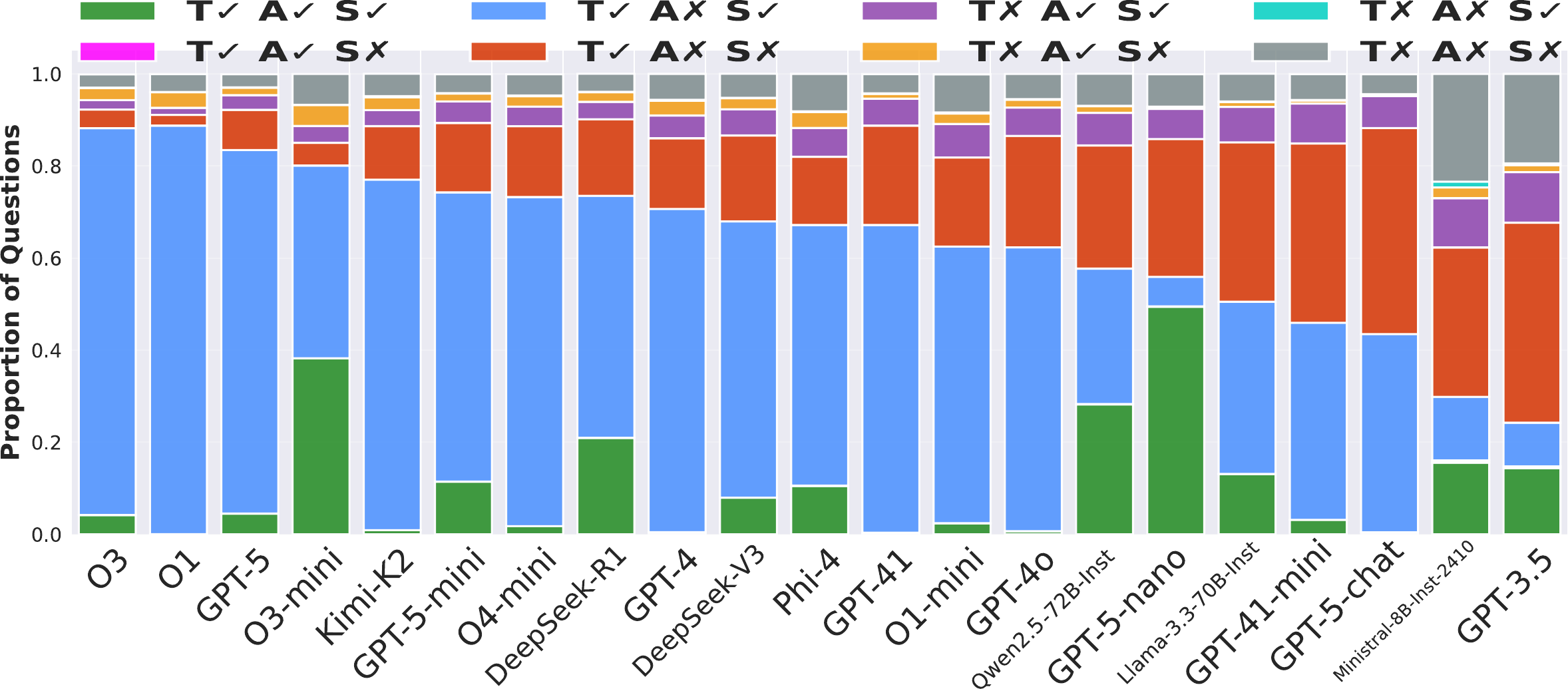}           
        \captionsetup{font=tiny} 
        \caption*{
          \parbox{\textwidth}{
            \centering
            \textbf{MMLU}
            }
        }
    \end{minipage}
        \captionsetup{font=normalsize}
        \caption{Eight dialectical reasoning patterns of different LLMs ({Notations:} T: Thesis, A: Antithesis, \& S: Synthesis. E.g.: {T\xmark}{A\cmark}{S\cmark} presents the ratio of times when T is incorrect, while A and S are correct.)}
        \label{fig:gsm_mmlu:8permutations}
\end{figure*}

\soheilhead{Static Accuracy vs.\ Reasoning Robustness}
Similar $p_T$ can mask \emph{fragile} reasoning trajectories. For instance on GSM8K, models that appear comparable at $p_T$ (e.g., GPT‑5‑chat and O3) \emph{diverge} at $p_S$—with GPT‑5‑chat ranking near the bottom at synthesis.
This discrepancy underscores a critical insight: high static correctness do not reflect genuine reasoning capabilities and conventional evaluation methods can \emph{overstate} model's reasoning robustness and quality when the generative path exhibits stability issues or \emph{pattern replay behavior}. The performance drops may hint at underlying issues in models training, though diagnosing these is beyond our current scope.

\begin{figure*}[tb]
\centering
\includegraphics[width=\textwidth]{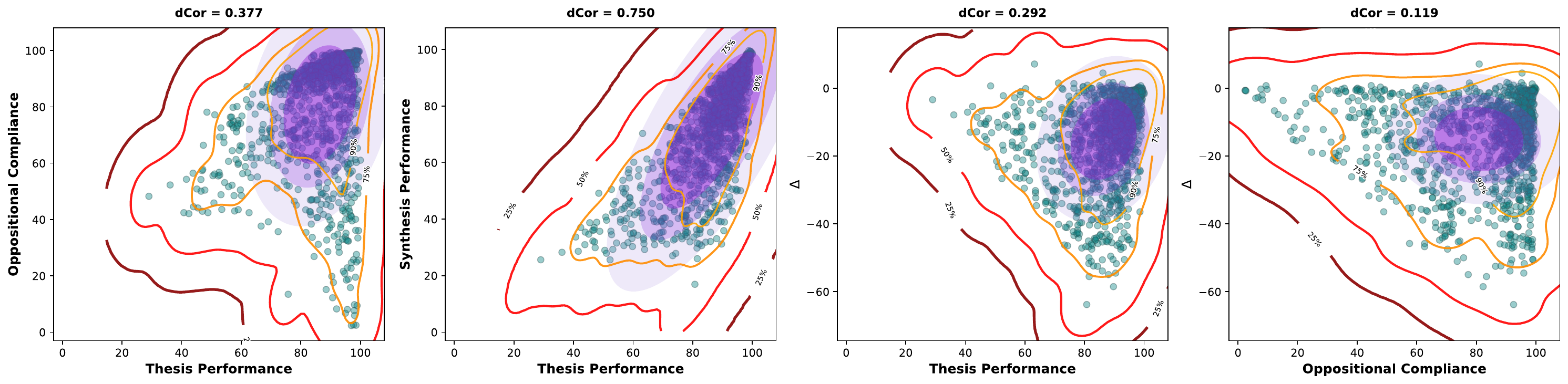}   
    \captionsetup{font=normalsize}
\caption{
Distance correlation analysis of dialectical reasoning in all MMLU sub-tests. Each teal circle represents one model-subject combination (21$\times$57). {Contours:} (1) \textit{Purple/pink confidence ellipses} show three dependence regions at 50\%, 75\%, and 95\% confidence levels with darker colors indicating higher confidence areas; (2) \textit{Orange/red density contour lines} marked with percentile labels (25\%, 50\%, 75\%, 90\%) represent data concentration levels, progressing from dark red (25th \%tile, highest density core) to light orange (90th \%tile, outer data boundary). The ellipses reveal the shape and orientation of dependencies, while the labeled density lines quantify data distribution patterns.}
\label{fig:distance_correlation:overall:1x4}
\end{figure*}

\soheilhead{DS vs. Synthesis Scoring}
$p_S$ is intuitive but can over-credit trajectories that rarely produce opposition. DS modulates $p_S$ by OC to better separate \emph{engaged integration} from \emph{replay-like} behavior. On GSM8K (Table~\ref{table:eval:overall}), GPT‑5‑nano attains respectable $p_S$ (84.4) yet exhibits very low OC (15.5\%), consistent with replay tendencies; GPT‑5 shows substantially higher OC (81.4\%), and DS reflects this difference (77.5 vs.\ 63.0). 
The eight permutations of thesis (T), antithesis (A), and synthesis (S), along with their occurrence ratios in Figure~\ref{fig:gsm_mmlu:8permutations} in Fig.~\ref{fig:gsm_mmlu:8permutations} further illustrate the pattern: frequent \texttt{T\cmark\,A\cmark\,S\cmark} agreement sequences align with replay-like trajectories, whereas higher OC ratios indicate consistent opposition production that DS rewards \emph{when} synthesis integrates successfully.


\soheilhead{$\Delta$ as an Adaptation Signal}
An interesting finding is that across models and benchmarks (Table~\ref{table:eval:overall}), $\Delta$ is negative on average, indicating limited refinement at synthesis—consistent with \emph{fragile} or \emph{replay-like} trajectories. At the same time, Fig.~\ref{fig:gsm_mmlu:8permutations} shows recurring \texttt{T\xmark\,A\cmark\,S\cmark} episodes (notably in several MMLU subtopics), confirming that integrative gains \emph{do occur}, albeit not dominantly. Reading $p_S$, OC, DS, and $\Delta$ together offers interpretable indicators of whether a model tends toward \emph{stable integration} or \emph{pattern replay} (check Appendix~\ref{app:mmlu:full:drix} for more details).

\soheilhead{Correlation Analysis}
To provide a complementary view, we apply distance correlation analysis~\cite{szekely2007measuring} to various metrics across all MMLU sub-topics. Figure~\ref{fig:distance_correlation:overall:1x4} shows several notable patterns.
First, in general, \emph{thesis correctness} ($p_T$) is \emph{weakly} related to both \emph{opposition production} (OC) and \emph{thesis$\!\to\!$synthesis change} ($\Delta$): standalone accuracy says little about whether a model {produces} opposing antitheses or {refines} its stance at synthesis. 
Similarly, OC and $\Delta$ are weakly linked, though low OC often leads to low $\Delta$, suggesting weak antitheses result in minimal deviation from the thesis.
The $p_T$–$p_S$ relationship is stronger yet \emph{inconsistent and non‑linear}: models with similar thesis scores often produce \emph{divergent} syntheses. 
Combined with the thesis–$\Delta$ trend, this shows that while high thesis scores don’t imply strong reasoning, low thesis scores reliably signal poor dialectical performance.
In short, models that perform poorly on conventional accuracy-based evaluations often lack deeper reasoning—but high standalone answer accuracy doesn’t guarantee genuine reasoning ability either. These findings underscore our motivation to move beyond static correctness and toward evaluating reasoning through structured, dynamic processes. 
Appendix~\ref{sec:app:correlation} provides further details on distance calculations and extended correlation analysis.


\begin{figure*}[t]
\centering
\begin{minipage}[b]{0.24\textwidth}
    \centering
    \includegraphics[width=\linewidth]{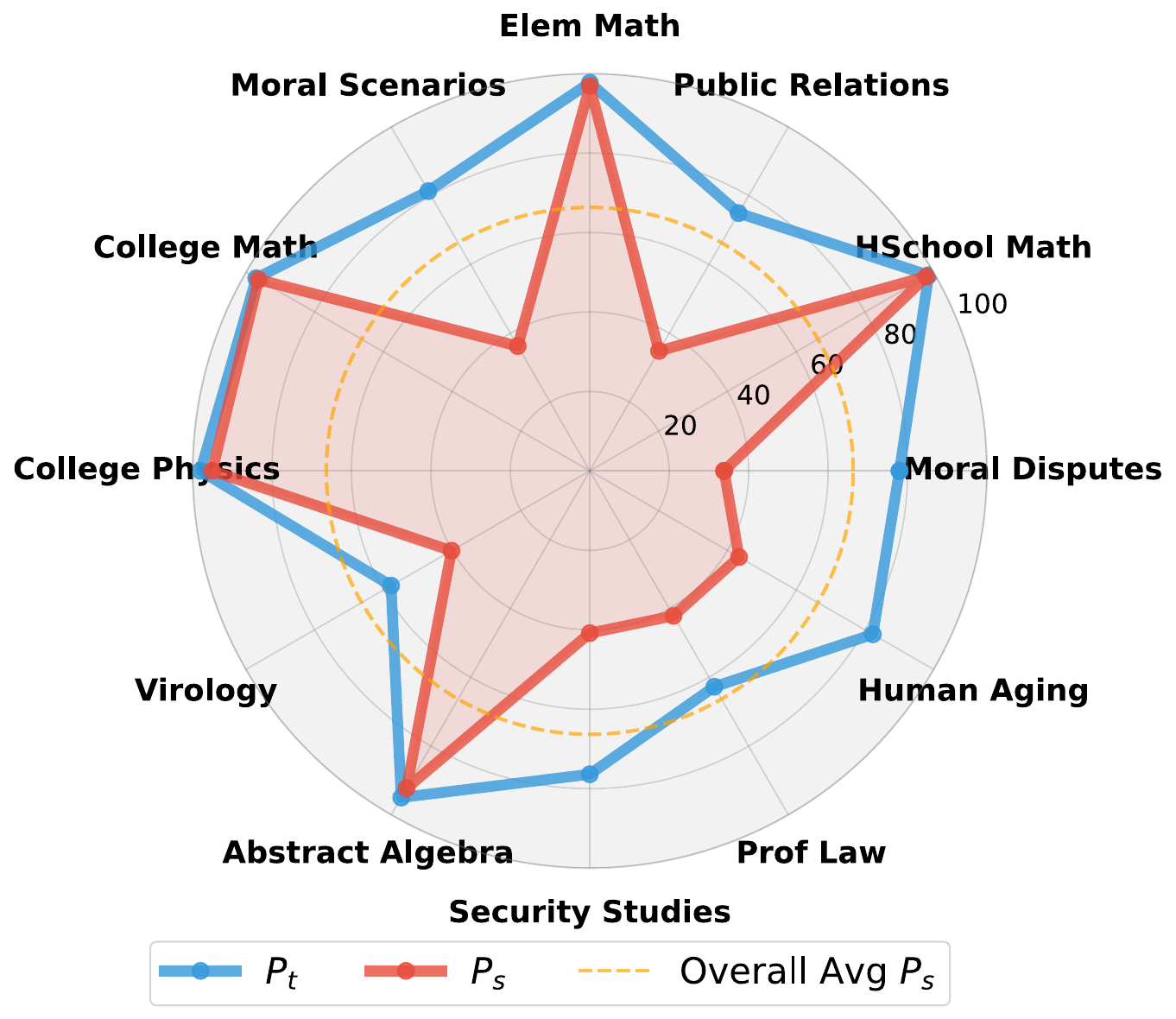}
    \captionsetup{font=tiny} 
    \caption*{
      \parbox{\textwidth}{
        \centering
        GPT-5-nano
      }
    }
    \captionsetup{font=normalsize}
\end{minipage}
\begin{minipage}[b]{0.24\textwidth}
    \includegraphics[width=\linewidth]{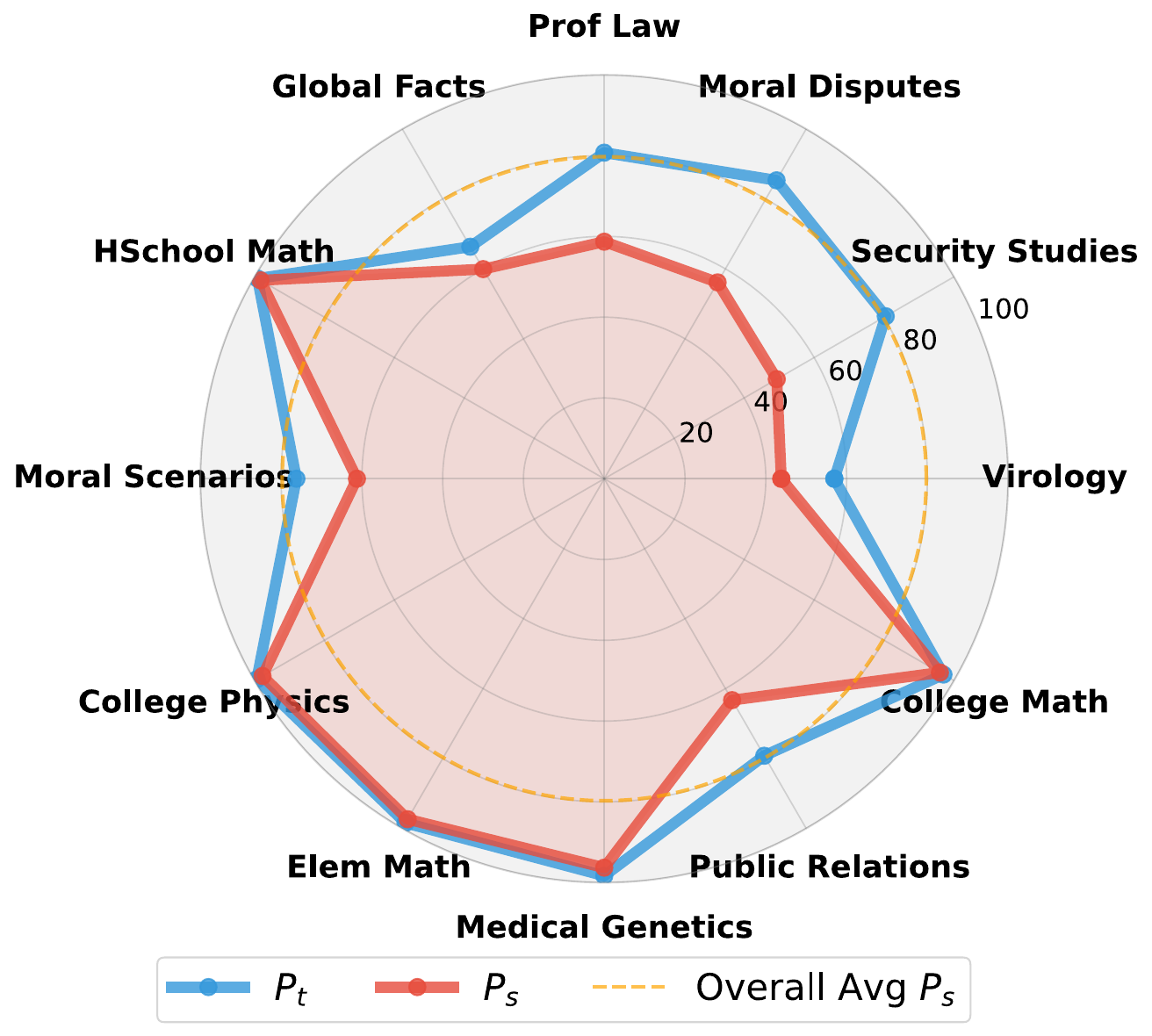}
    \captionsetup{font=tiny} 
    \caption*{
      \parbox{\textwidth}{
        \centering
        DeepSeek-R1
      }
    }
    \captionsetup{font=normalsize}
\end{minipage}
\begin{minipage}[b]{0.24\textwidth}
    \includegraphics[width=\linewidth]{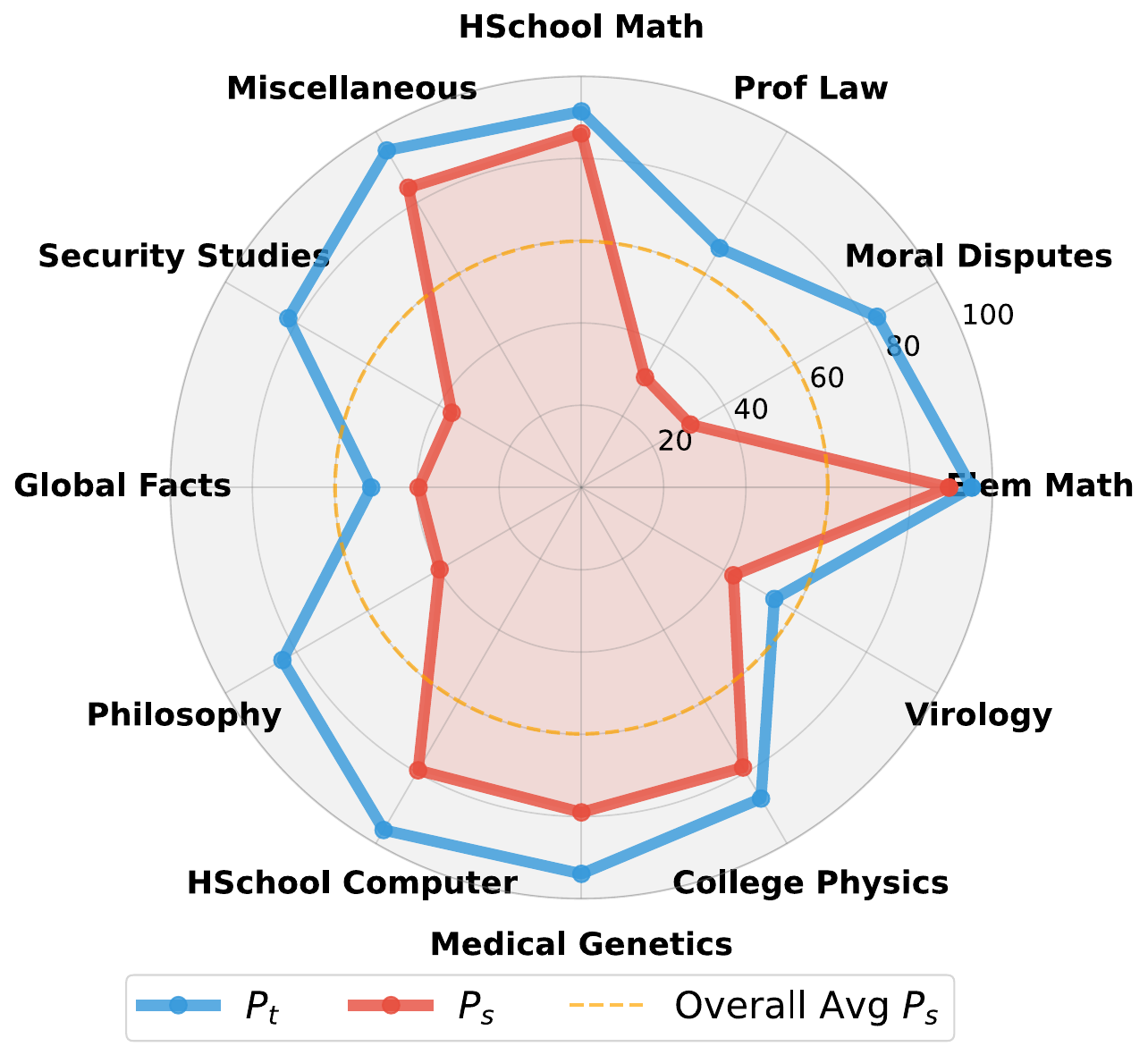}
    \captionsetup{font=tiny} 
    \caption*{
      \parbox{\textwidth}{
        \centering
        Llama3.3-70B-Inst
      }
    }
    \captionsetup{font=normalsize}
\end{minipage}
\begin{minipage}[b]{0.24\textwidth}
    \includegraphics[width=\linewidth]{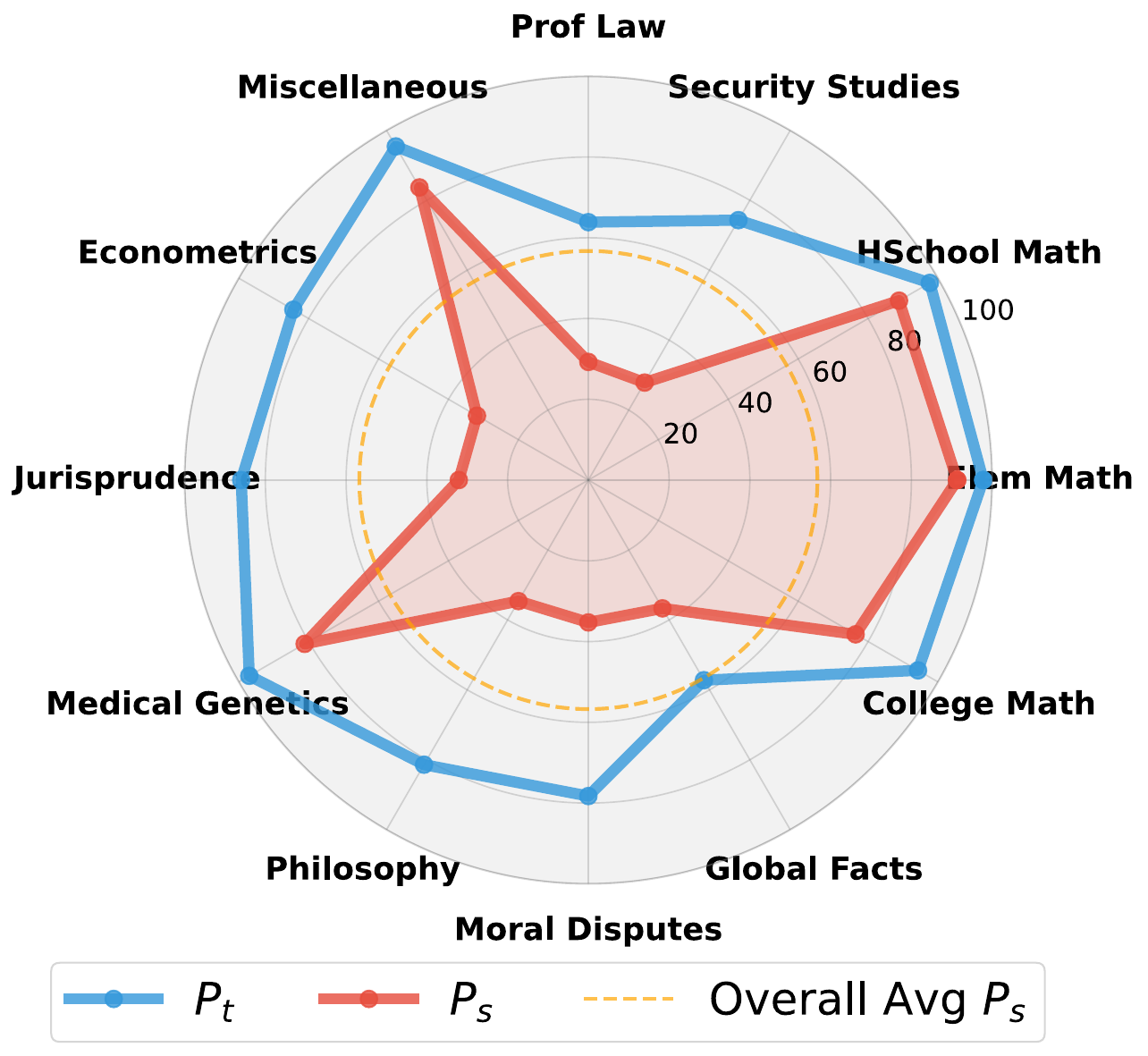}
    \captionsetup{font=tiny} 
    \caption*{
      \parbox{\textwidth}{
        \centering
        GPT-4.1-mini
      }
    }
    \captionsetup{font=normalsize}
\end{minipage}
\caption{Samples of reasoning performance ($p_S$, red regions) of models on diff. MMLU topics}
\label{fig:radar:mmlu:topics}
\end{figure*}
\textbf{General Reasoning Skill or Topic-Oriented?}
Results indicate that models reasoning is strongly topic-dependent rather than a uniform, general capability. Model rankings shift with the benchmark (e.g., Llama3.3-70B-Instruct performs well on GSM but is mixed on MMLU, whereas O1 shows the opposite trend), indicating topic-specific strengths. Figure~\ref{fig:radar:mmlu:topics} makes this explicit: \(p_S\) varies widely across domains in MMLU. 
For instance, Llama3.3-70B-Instruct attains high \(p_S\) in Elementary Math yet lags markedly in normative areas such as Moral Disputes (and related topics like Security Studies); DeepSeek-R1 peaks in quantitative subjects (e.g., mathematics and physics) but similarly weakens on normative domains. These patterns may suggest that what looks like “reasoning ability” often reflects uneven exposure to domain-specific structures during training—an imprint of imbalanced data distributions and topic-specific regularities seen during learning—rather than a genuinely general skill. 

\begin{figure*}[!tp]
    \centering
    \begin{minipage}[b]{0.325\textwidth}
        \centering
        \includegraphics[trim=0.25cm 0cm 0cm 0cm, clip,width=\textwidth]{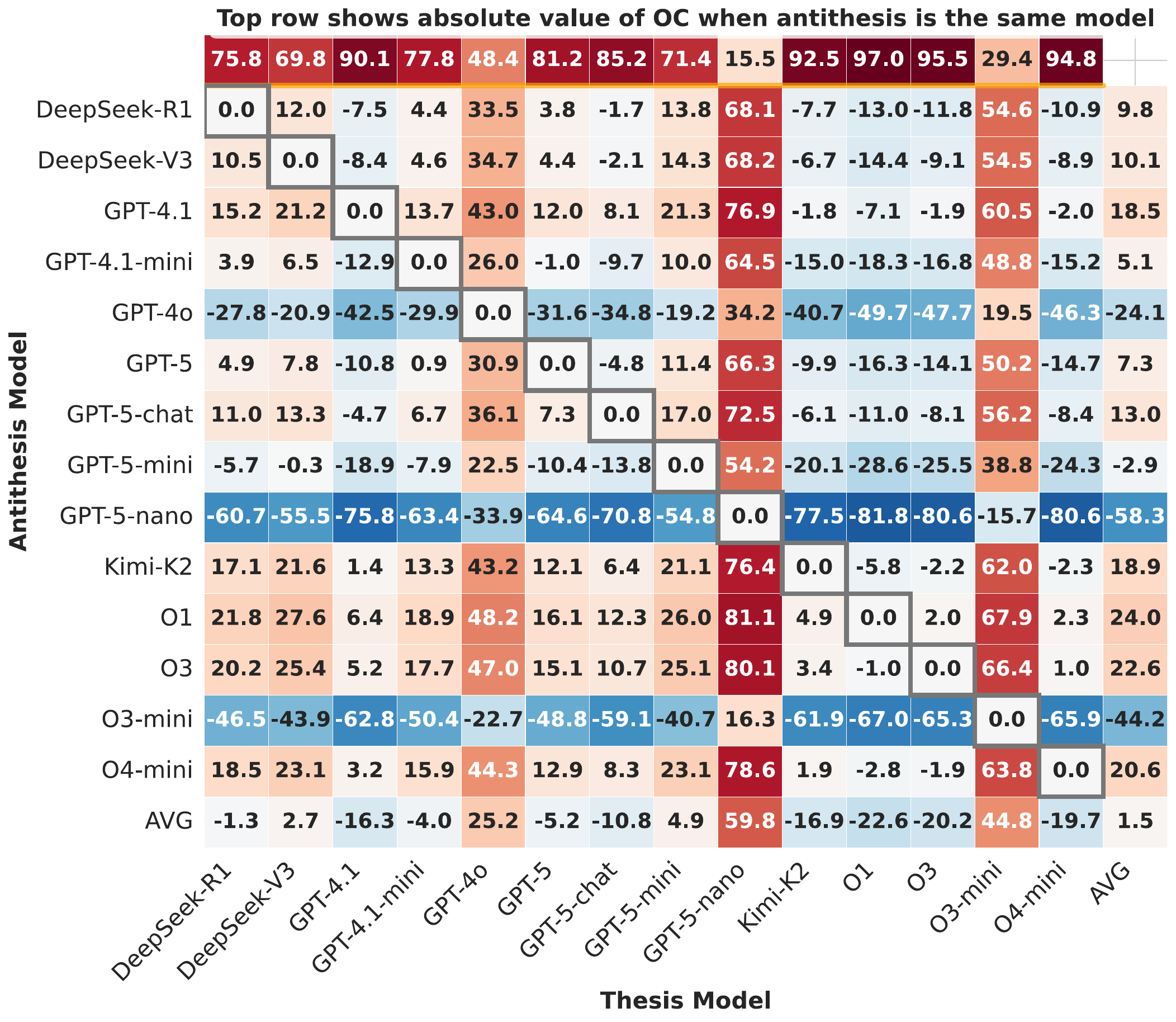}
    \end{minipage}   
    \begin{minipage}[b]{0.325\textwidth}
        \centering
        \includegraphics[trim=0.4cm 0cm 0cm 0cm, clip,width=\textwidth]{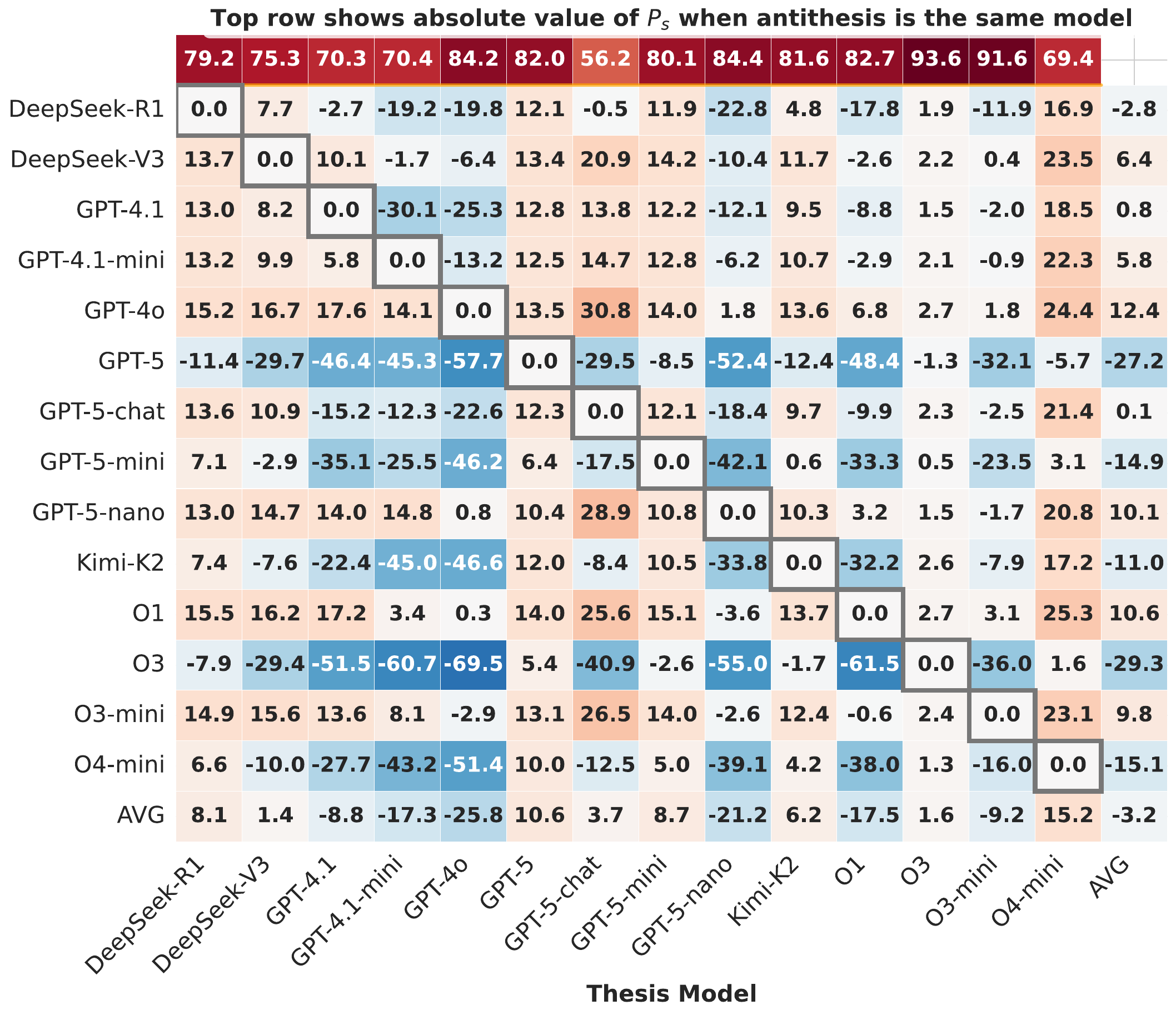}    
    \end{minipage}
    \begin{minipage}[b]{0.325\textwidth}
        \centering
        \includegraphics[trim=1.1cm 0cm 0cm 0cm, clip,width=\textwidth]{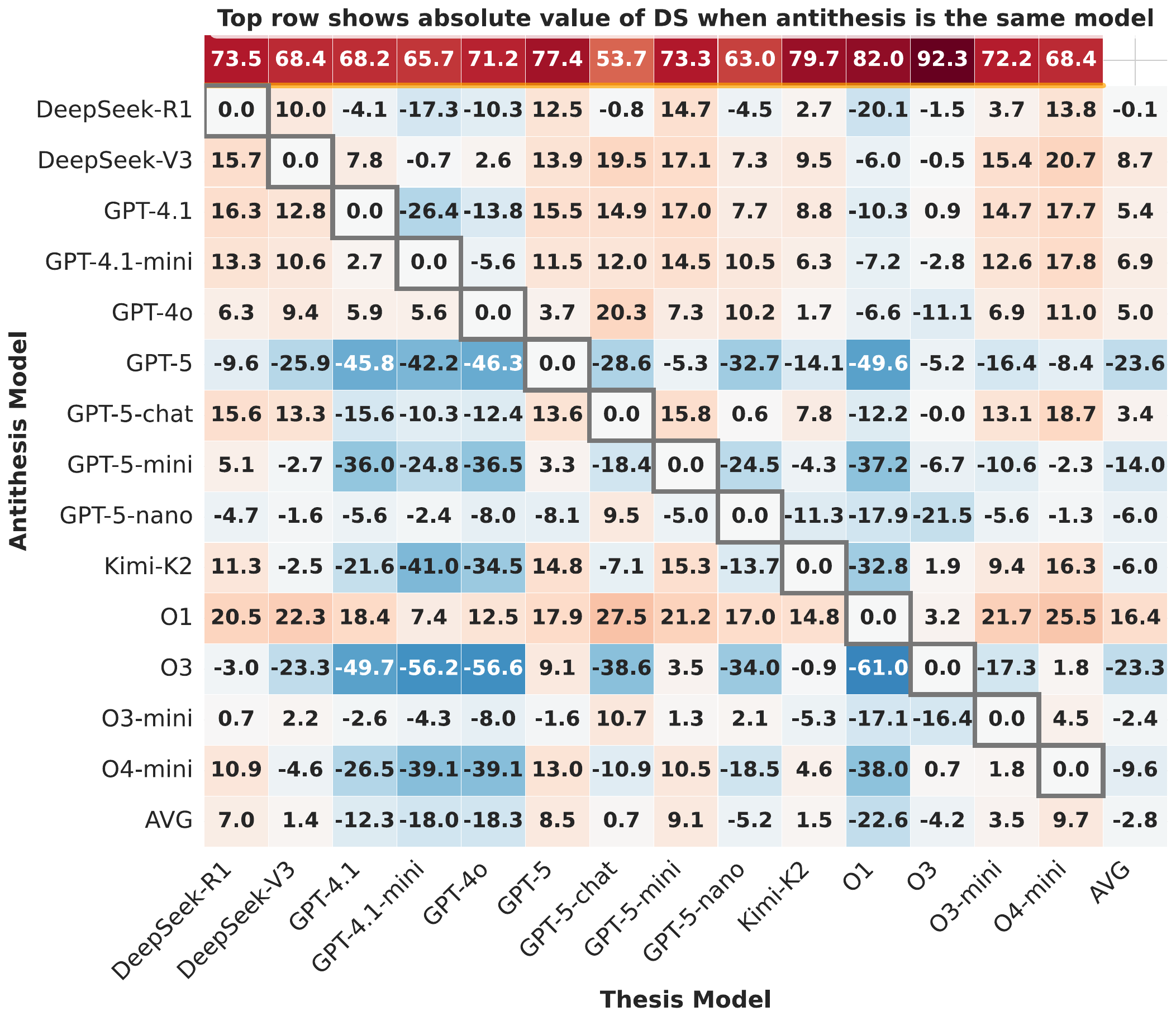}    
    \end{minipage}
    \caption{left: OC, middle: $p_S$, right: DS, cross-model dialectics for GSM. Numbers in cells show $($cross-X $-$ self-X$)$ values, where X is OC, $p_S$, and DS, respectively, from left to right. The top row in each graph shows the self-X values.}
    \label{fig:cross-model:gsm}
\end{figure*}

\soheilhead{Model Size and Vintage Effects}
Process‑level signals in Table~\ref{table:eval:overall} often reveal differences that $p_T$ obscures. Several large–small pairs (e.g., GPT‑4.1 vs.\ GPT‑4.1‑mini; GPT‑5 vs.\ GPT‑5‑mini; O1 vs.\ O1‑mini) show substantially wider gaps in $p_S$, DS, and $\Delta$ than in $p_T$. Likewise, predecessor–successor contrasts appear in synthesis behavior (e.g., GPT‑4 vs.\ GPT‑4o/GPT‑4.1 on GSM8K), suggesting that changes across model generations and different training strategies affects the {stability and integrative quality} of reasoning—even when thesis correctness remains high. 

\subsection{Cross-Model Dialectics}
\label{sec:eval:cross-model}
In a multi-agent ecosystem, it will be common to see agents powered by different base models to collaborate on shared tasks. In such settings, a dialectical reasoning evaluation that captures the communicative dimension of reasoning becomes even more critical. So, here, we assess how a model's reasoning performance is influenced when the antithesis is generated by a different model—potentially with a distinct internal structure. 

\soheilhead{Setting}
To that end, we configure Agent B (Figure~\ref{fig:sieve:pipeline}) to use a different LLM for generating the antithesis. We then evaluate the reasoning performance of the primary (thesis) model. 
For brevity, we report GSM results here and differ MMLU to Appendix~\ref{app:cross-model:mmlu}. These experiments cover 14 models.

\soheilhead{Cross-Model OC}
In cross‑model scenarios we observe a clear and consistent pattern: a model’s \emph{self‑OC} 
strongly predicts how much it can raise (or lower) the \emph{cross-OC} of other models when acting as the antithesis generator. As the left plot of Figure~\ref{fig:cross-model:gsm} shows, models with higher self‑OC tend to increase partners’ cross‑OC, while models with low self‑OC reduce it.
For example, the three lowest self‑OC models (GPT‑5‑nano, O3‑mini, GPT‑4o) decrease cross‑OC for all other models with higher self‑OC. In contrast, high self‑OC models (O1, O3, O4‑mini) consistently improve cross‑OC of models whose self‑OC is lower. Moreover, the averaged cross‑model improvement ranking largely mirrors the self‑OC ranking. These observations indicate that self‑OC is a strong proxy for a model’s general antithesis‑generation competence that transfers across thesis distributions.

\soheilhead{Cross-Model Reasoning Performance}
The middle plot of Figure~\ref{fig:cross-model:gsm} shows changes of $p_S$ compared to normal self-dialectical scenarios. Many models show notable reasoning gains when paired with a different antithesis model. For instance, GPT‑5 improves across all pairings we tested, with gains ranging from $+5.4$ to $+14$ points in $p_S$. Similar patterns appear for DeepSeek‑R1 and O4‑mini in most cross‑model settings. 
Consistent with the $p_S$ patterns, $DS$ (right plot of Figure~\ref{fig:cross-model:gsm}) generally rises with better pairings but shows a floor effect when antitheses are too agreeable or weak: very low‑OC generators (GPT‑5‑nano, O3‑mini, GPT‑4o) fail to provide high quality antithesis, in turn, limiting $DS$. Overall, synthesis tends to benefit from \emph{diverse, substantive} antitheses, whereas low‑OC pairings align with \emph{pattern replay behavior}.

\soheilhead{Key Takeaway}
At first glance, these improvements might make it tempting to conclude that models become “better reasoners” when exposed to diverse antitheses. However, the variability across pairings invites a more cautious interpretation: could what is called “reasoning” in LLMs be less a general, stable capability and more a context‑sensitive skill shaped by input structure? Prior work has raised similar concerns, questioning whether LLMs genuinely reason or merely mimic reasoning through statistical pattern matching~\cite{dziri2023faith, kambhampati2024can, nezhurina2024alice, mccoy2023embers}. The gains we observe here may reflect how certain antithesis forms provide structural signals that align with token‑level patterns the model has internalized, rather than signaling a universal reasoning ability. If mere structural familiarity were the dominant factor, one might expect a model to perform best when paired with its own antithesis—yet this is not necessarily the case. Self‑generated antitheses tend to resemble the thesis in tone and structure, reducing the contrast needed for producing effective synthesis. In contrast, cross‑model antitheses introduce greater diversity—different token rhythms, alternative rhetorical styles—that can create stronger oppositional signals and more effective support for synthesis. These patterns do not settle the debate, but they add weight to an existing view~\cite{jiang2024peek, schaeffer2023emergent, razeghi2022impact}: what looks like reasoning may, in practice, be a skill that thrives on structural variety rather than a general cognitive ability.

\section{Brief Discussion and Limitations}
\soheilhead{SIEV as a General Approach}
It is important to note that SIEV is not a benchmark itself, but a dialectical approach to benchmark models. Rather than specifying exact questions to be solved (as benchmarks do), it defines how the assessment process should be conducted. In other words, SIEV can be applied to any benchmark and is not tied to a specific one. This flexibility allows SIEV to serve as a general approach for assessing the reasoning capabilities of LLMs across diverse settings and topics.





\textbf{Scope of Benchmarks:}
Our analysis focuses on GSM8K and MMLU—benchmarks that are both widely used and largely saturated. Although SIEV exposes substantial reasoning variability on these datasets, it remains to be seen how these findings generalize to emerging benchmarks, multimodal settings, or tasks that demand long‑horizon planning or domain‑specific symbolic reasoning.

\textbf{Opposition Quality:}
Opposition Compliance (OC) measures whether a model produces an antithesis that \emph{opposes} the thesis, but does not evaluate the \emph{semantic quality} of that opposition. A generated antithesis may be formally opposing yet shallow or irrelevant. Extending OC to capture degrees or categories of opposition—while preserving automation—remains an important direction.

\textbf{Synthesis Quality and Evaluation Granularity:}
SIEV evaluates synthesis via correctness ($p_S$) and DS. However, synthesis quality can be multidimensional: integration can be logically coherent yet factually incorrect, or vice versa. More fine‑grained assessments—e.g., evaluating causal structure, logical entailment, or self‑consistency—could yield richer insights.

\textbf{Absence of Human‑Judged Reasoning Traces:}
SIEV intentionally avoids reliance on manual annotations, enabling scalable evaluation. Yet some aspects of reasoning—such as whether an antithesis is conceptually meaningful, or whether a synthesis exhibits genuine insight—may require expert human judgment. Human‑in‑the‑loop extensions could validate and refine SIEV’s automated signals.

Overall, SIEV offers a step toward process‑aware reasoning evaluation, but further work is required to broaden coverage, enrich reasoning signals, and incorporate both human judgment and robustness analyses.

\section{Concluding Remarks}
Evaluating LLM reasoning requires looking beyond whether a model arrives at the correct answer. As our study shows, correctness and even polished CoT traces often conceal the nature of the underlying reasoning—whether it reflects a stable, integrative process or a replay of familiar training patterns. 
SIEV provides a structured way to make these differences visible. 
By eliciting thesis–antithesis–synthesis interactions, it captures signals such as opposition production, synthesis quality, and directional refinement that conventional accuracy obscures. Using these process-level indicators, we find that models with similar conventional accuracy scores can exhibit sharply different reasoning behaviors: some consistently integrate competing perspectives, while others default toward replay-like trajectories.

These insights highlight the importance of assessing reasoning as a dynamic trajectory rather than a static final output. As LLMs increasingly participate in scientific analysis, education, and multi-agent collaboration, such process-aware evaluation becomes essential for understanding not only \emph{what} models conclude but also \emph{how} and \emph{why} they arrive there. SIEV contributes toward this goal by offering a practical, transparent means to characterize the robustness and structure of LLM reasoning.

\section*{Impact Statement}


This paper presents work whose goal is to advance the field of Machine
Learning. There are many potential societal consequences of our work, none
which we feel must be specifically highlighted here.


\bibliography{ref}
\bibliographystyle{icml2026}

\newpage
\appendix
\onecolumn

\section{Prompt Specifications: Thesis--Antithesis--Synthesis Pattern}
\label{app:prompts}
The general prompt and specific parameters used for MMLU and GSM tasks are as follows. 

{
\footnotesize
\begin{infobox}{GENERAL FORMAT (Benchmark-Agnostic)}
\begin{thesisbox}{Stage 1: THESIS PROMPT}
You have extensive world knowledge and problem solving ability and great in solving \texttt{\textless TASK TYPE\textgreater} questions.

You are tasked to solve the following \texttt{\textless TASK TYPE\textgreater} question.  

\vspace{5pt}
\texttt{--PROBLEM STATEMENT--}

\vspace{5pt}
To do so, you need to provide the answer and explain the reasoning behind that.  
Your answer must follow the following format and must include the terms \texttt{\_reasoning\_:} and \texttt{\_final\_answer\_:}:

\begin{verbatim}
_reasoning_:
<Your reasoning comes here>
_final_answer_: (X)
\end{verbatim}

where X must be \texttt{\textless DOMAIN-SPECIFIC FORMAT\textgreater}.

\texttt{\textless examples of a valid response\textgreater}  

\texttt{\textless examples of an invalid response\textgreater}

\vspace{5pt}

Considering these instructions, answer the mentioned question.
\end{thesisbox}

\vspace{0.8em}

\begin{antithesisbox}{Stage 2: ANTITHESIS  PROMPT}
You have extensive world knowledge and problem solving ability and great in solving \texttt{\textless TASK TYPE\textgreater} questions.

Your task is to provide a contrasting perspective on a provided solution for a given \texttt{\textless TASK TYPE\textgreater} question.

\vspace{5pt}
Provide direct, precise criticism to the given answer. 
In particular, challenge strategy, decisions, reasoning provided in the given answer even if they are valid, and offer great antithesis accordingly. 
Your antithesis should provide another way to look and solve the given problem while opposing the entirety of the solutions provided.  
consequently, determine what can be a better final answer.

Ensure your responses are concise and to the point without being unnecessarily long.

\vspace{5pt}
\textbf{Context}: 
The question is: 

\texttt{--PROBLEM STATEMENT--}  

The given answer is:  

\texttt{--THESIS STATEMENT--}

\vspace{5pt}
Criticize the given answer and reasoning, clearly explaining your reasoning and antithesis. Your response must follow this format:

\begin{verbatim}
_reasoning_:
\end{verbatim}
[Your reasoning here and why you think X is a better answer]
\begin{verbatim}
_final_answer_: (X)
\end{verbatim}

where X is \texttt{\textless DOMAIN-SPECIFIC FORMAT\textgreater}.

\vspace{5pt}
\texttt{\textless example of a valid response\textgreater}  

\texttt{\textless example of an invalid response\textgreater}

\vspace{5pt}
Now, review the provided solution, offer great criticism and antithesis and provide your response following the template.

\vspace{5pt}
Remember that you do NOT need to provide the correct final answer to the given question, 
your task is to professionally challenge the provided answer/reasoning EVEN if you think the answer is already correct! 
That means you should challenge it no matter what!

\end{antithesisbox}

\vspace{0.8em}

\begin{synthesisbox}{Stage 3: SYNTHESIS  PROMPT}
To help you on your task, I provide you with the responses of another agent who is observing your answers:

\texttt{\textless insert antithesis responses here\textgreater}

\vspace{5pt}
Now, given the arguments and comments provided, update your answer.  
Keep the same format as before when you are responding.

\end{synthesisbox}

\end{infobox}
}

{
\footnotesize
\begin{infobox}{MMLU PARAMETERS}
\texttt{\textless TASK TYPE\textgreater} = "multiple-choice"

\vspace{5pt}
\texttt{\textless DOMAIN-SPECIFIC FORMAT\textgreater} = "A, B, C, or D"
\vspace{5pt}
\end{infobox}


\begin{infobox}{GSM PARAMETERS}

\texttt{\textless TASK TYPE\textgreater} = "math"

\vspace{5pt}
\texttt{\textless DOMAIN-SPECIFIC FORMAT\textgreater} = "a number"
\vspace{5pt}
\end{infobox}
}

\section{Another Illustrative Example of SIEV Pipeline}
\label{app:siev:examples}
Figure~\ref{fig:siev:example:gpt5-v2} presents another illustrative example of the SIEV pipeline, showcasing a dialectical reasoning evaluation using the GPT-5 model and a conceptual physics question from the MMLU benchmark. Unlike Figure~\ref{fig:siev:example:o3-r1}, where models encountered invalid opposing views, here the GPT‑5 model faces a valid antithesis. The thesis is incorrect while the antithesis identifies the correct answer; however, the synthesis closely mirrors the thesis’s flawed rationale. 
In this example, GPT-5 model fails to recognize the valid opposing view and instead favors the thesis response as higher quality—ultimately producing an incorrect synthesis and invalid reasoning.


\newcommand{\Poblemxone}{
According to four-dimensional geometry, the angles of a triangle add to $180^\circ$:\\ A) always., B) sometimes., C) never., D) on planet Earth only.
}

\newcommand{\Answerxone}{
    \scriptsize
\textbf{Answer: B}
}

\newcommand{\ThesisOnexone}{
\textbf{Reasoning:}
In Euclidean geometry of any dimension, any three non-collinear points define a 2D plane, and a triangle lies entirely in that plane. Therefore, in 4-dimensional Euclidean space, the interior angles of a triangle still sum to $180^\circ$, just as in the plane.

\textbf{Why others are wrong:}
\begin{itemize}
    \item \textbf{B) sometimes:} This would be true in curved geometries (spherical or hyperbolic), where angle sums deviate from $180^\circ$. But the phrase ``four-dimensional geometry'' by itself (as Euclidean 4D space) does not change the triangle’s angle sum; it remains $180^\circ$ for all triangles.
    
    \item \textbf{C) never:} This contradicts Euclidean results; in flat 4D space the sum is always $180^\circ$.
    
    \item \textbf{D) on planet Earth only:} On Earth’s approximately spherical surface, spherical triangles have sums greater than $180^\circ$, not exactly $180^\circ$. Moreover, the statement ties it incorrectly to Earth rather than to the curvature of the surface.
\end{itemize}
\textbf{Answer: A}
}

\newcommand{\ThesisTwoxone}{

}

\newcommand{\AntiThesisOnexone}{
\textbf{Reasoning:}
The given answer smuggles in an unstated assumption: that ``four-dimensional geometry'' means flat Euclidean $\mathbb{R}^4$. That’s a narrow reading. Dimension does not determine angle sums; the metric/curvature does. In 4D, there are many canonical geometries—Euclidean (flat), spherical ($\mathbb{S}^4$), hyperbolic ($\mathbb{H}^4$), and pseudo-Riemannian (e.g., Minkowski spacetime). 

In flat $\mathbb{R}^4$, straight-line triangles are indeed planar with $180^\circ$ angle sum—but that’s only one case. In 4D curved spaces, geodesic triangles have angle sums $\neq 180^\circ$. Even within Euclidean $\mathbb{R}^4$, if ``triangle'' refers to geodesics on a 3D hypersurface (like $\mathbb{S}^3$ embedded in $\mathbb{R}^4$), the intrinsic triangle angle sum deviates from $180^\circ$. 

The question’s wording doesn’t specify Euclidean; thus ``always'' is overreaching. Conversely, ``never'' is false (flat 4D gives $180^\circ$), and ``on planet Earth only'' is irrelevant and incorrect. Therefore, ``sometimes'' best captures the dependence on curvature/metric, which is the real determinant—not the dimension.
\\
\textbf{Answer: B}
}

\newcommand{\SynthesisOnexone}{
\textbf{Reasoning:}
In Euclidean geometry of any dimension, any three non-collinear points determine a unique 2D plane, and a triangle lies entirely within that plane. Therefore, in Euclidean 4-dimensional space (the usual default when one says ``four-dimensional geometry'' without further qualifiers), the interior angles of a triangle always sum to $180^\circ$.

\textbf{Why the others are wrong:}
\begin{itemize}
    \item \textbf{B) sometimes:} ``Sometimes'' would apply if we were talking about non-Euclidean 4D geometries (e.g., spherical or hyperbolic 4-manifolds) or triangles constrained to curved hypersurfaces, where the sum can differ from $180^\circ$. But absent any mention of curvature, the standard interpretation is Euclidean 4-space, in which it is always $180^\circ$.
    
    \item \textbf{C) never:} This is false because in flat (Euclidean) 4D space the sum is $180^\circ$ for every triangle.
    
    \item \textbf{D) on planet Earth only:} On Earth’s approximately spherical surface, geodesic triangles actually have sums greater than $180^\circ$, so this is incorrect, and the statement is unrelated to the dimension issue.
\end{itemize}
\textbf{Answer: A}
}

\begin{figure*}[th]
\centering
\resizebox{\textwidth}{!}{%
\begin{tikzpicture}[x=1cm, y=1cm]

\node[qCard, text width=0.95\textwidth] (Q) at (-1,0) {
      \scriptsize
  \justify
  \Poblemxone
};

\node[tCard, below=2.7cm of Q.south west, anchor=west, text width=0.95\textwidth] (T) {
      \scriptsize
  \justify  
  \ThesisOnexone
};

\node[aCard, below=2cm of T.south west, anchor=west, text width=0.95\textwidth] (A) {
    \scriptsize
  \justify
  \AntiThesisOnexone
};

\node[sCard, below=4.7cm of A.east, anchor=east, text width=0.95\textwidth, align=justify] (S) {
    \scriptsize
    \justify
    \SynthesisOnexone
};

\node[coordinate, below=.3cm of S.south] (Sbottom) {};
\node[coordinate, below=.3cm of S2.south] (Sbottom2) {};

\begin{pgfonlayer}{background}
  \node[draw=black!40, fill=gray!5, dashed, rounded corners, fit=(T)(A)(S)(Sbottom), inner sep=0.15cm] (back1) {};
\end{pgfonlayer}

\node at ($(Sbottom.south) + (0,+2pt)$) {\textbf{GPT-5 Model}};

\node[qPill, anchor=north west] at ($(Q.north west)+(0pt,7pt)$) {\faQuestionCircle\; Question};
\node[tPill, anchor=north west] at ($(T.north west)+(0pt,9pt)$) {\faLightbulb\; Thesis};
\node[aPill, anchor=north west] at ($(A.north west)+(0pt,9pt)$) {\faBolt\; Antithesis};
\node[sPill, anchor=north west] at ($(S.north west)+(0pt,9pt)$) {\faBrain\; Synthesis};

\node[chip, anchor=north east] at ($(Q.south east)+(-6pt,16pt)$) {\Answerxone};
\node[chip, anchor=north east] at ($(T.south east)+(-6pt,16pt)$) {\xmark};
\node[chip, anchor=north east] at ($(A.south east)+(-6pt,16pt)$) {\cmark};
\node[chip, anchor=north east] at ($(S.south east)+(-6pt,16pt)$) {\xmark};



\end{tikzpicture}%
}
\caption{An illustration of dialectical reasoning evaluation in SIEV using the GPT-5 model and a question from the conceptual physics section of the MMLU benchmark. In this example, the antithesis provided by GPT-5 model correctly identifies the answer, but even after observing this different angle, the model produces a synthesis that closely mirrors the flawed reasoning of the incorrect thesis.}
\label{fig:siev:example:gpt5-v2}
\end{figure*}

\section{Cross-Model Evaluations with MMLU}
\label{app:cross-model:mmlu}
Since the MMLU benchmark is extensive, we select five representative topics to reduce the experimental load while maintaining coverage of the benchmark’s diversity. These topics effectively capture the performance trends of different models across all MMLU categories. Specifically, we chose: (1) US foreign policy, (2) Management, (3) Computer security, (4) Public relations, and (5) Business ethics.

\begin{figure*}[t]
    \centering
    \begin{minipage}[b]{0.48\textwidth}
        \centering
        \includegraphics[trim=0.0cm 0cm 0cm 0cm, clip,width=\textwidth]{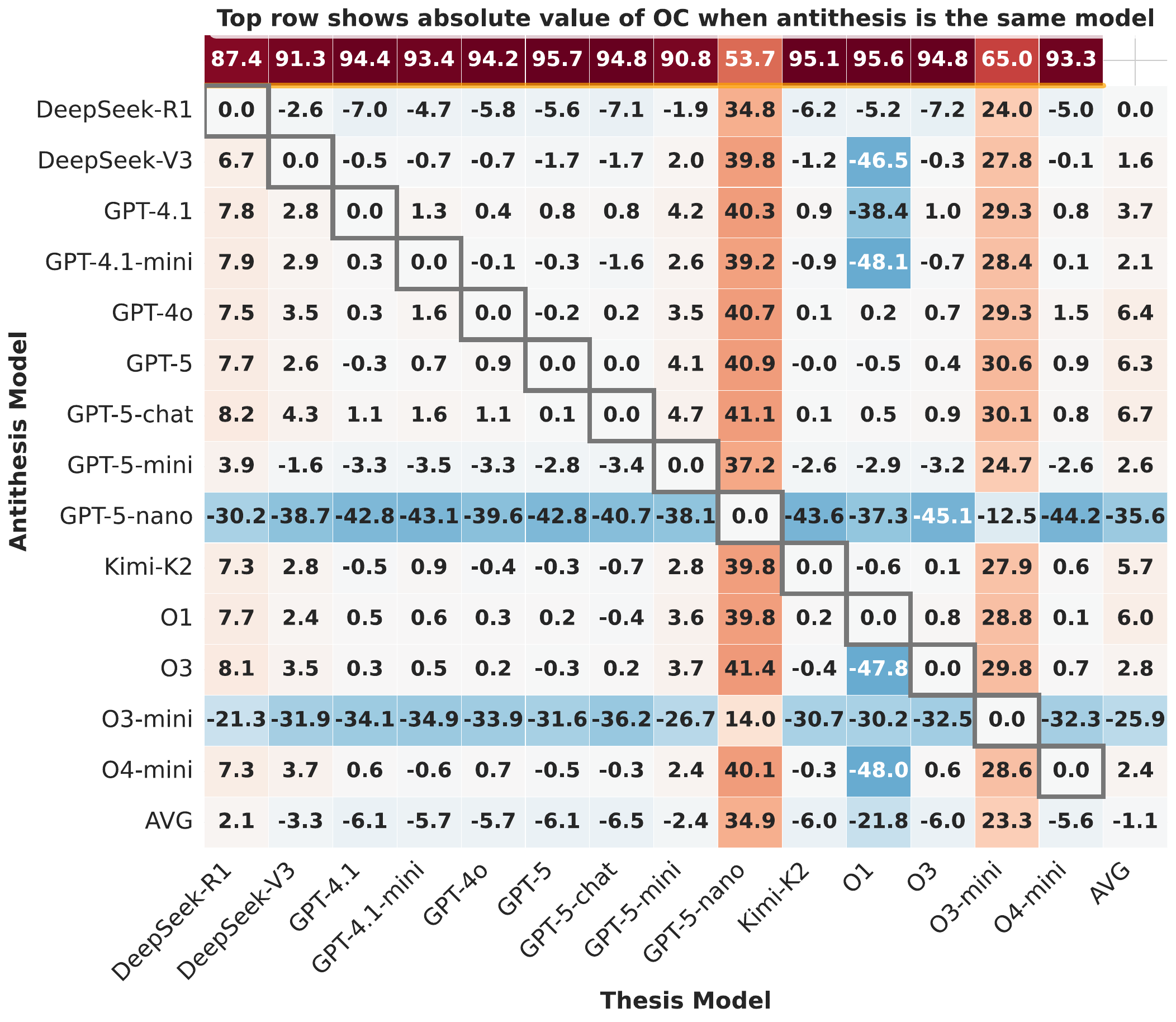}
    \end{minipage}   
    \begin{minipage}[b]{0.48\textwidth}
        \centering
        \includegraphics[trim=0.0cm 0cm 0cm 0cm, clip,width=\textwidth]{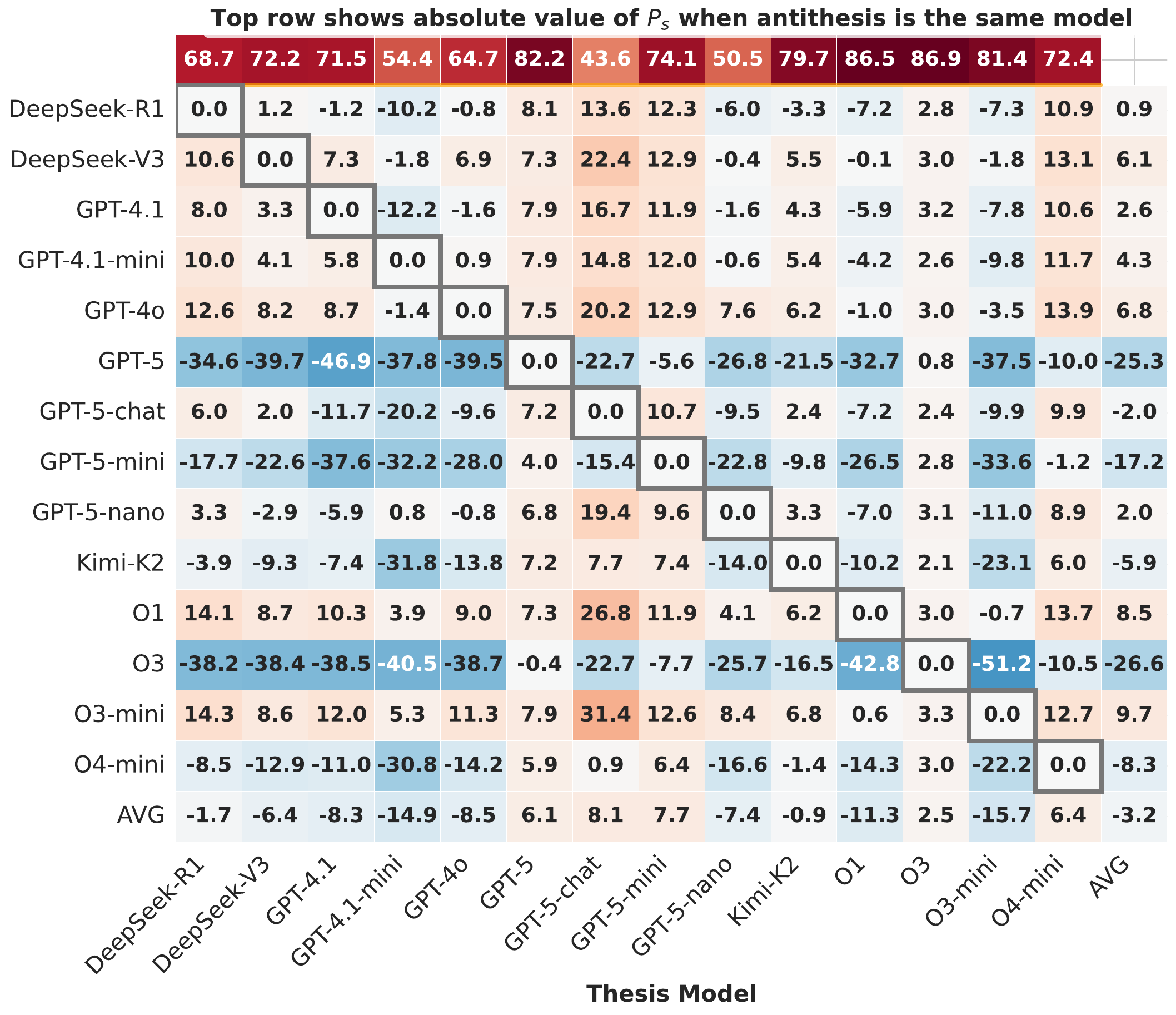}    
    \end{minipage}
    \begin{minipage}[b]{0.48\textwidth}
        \centering
        \includegraphics[trim=0cm 0cm 0cm 0cm, clip,width=\textwidth]{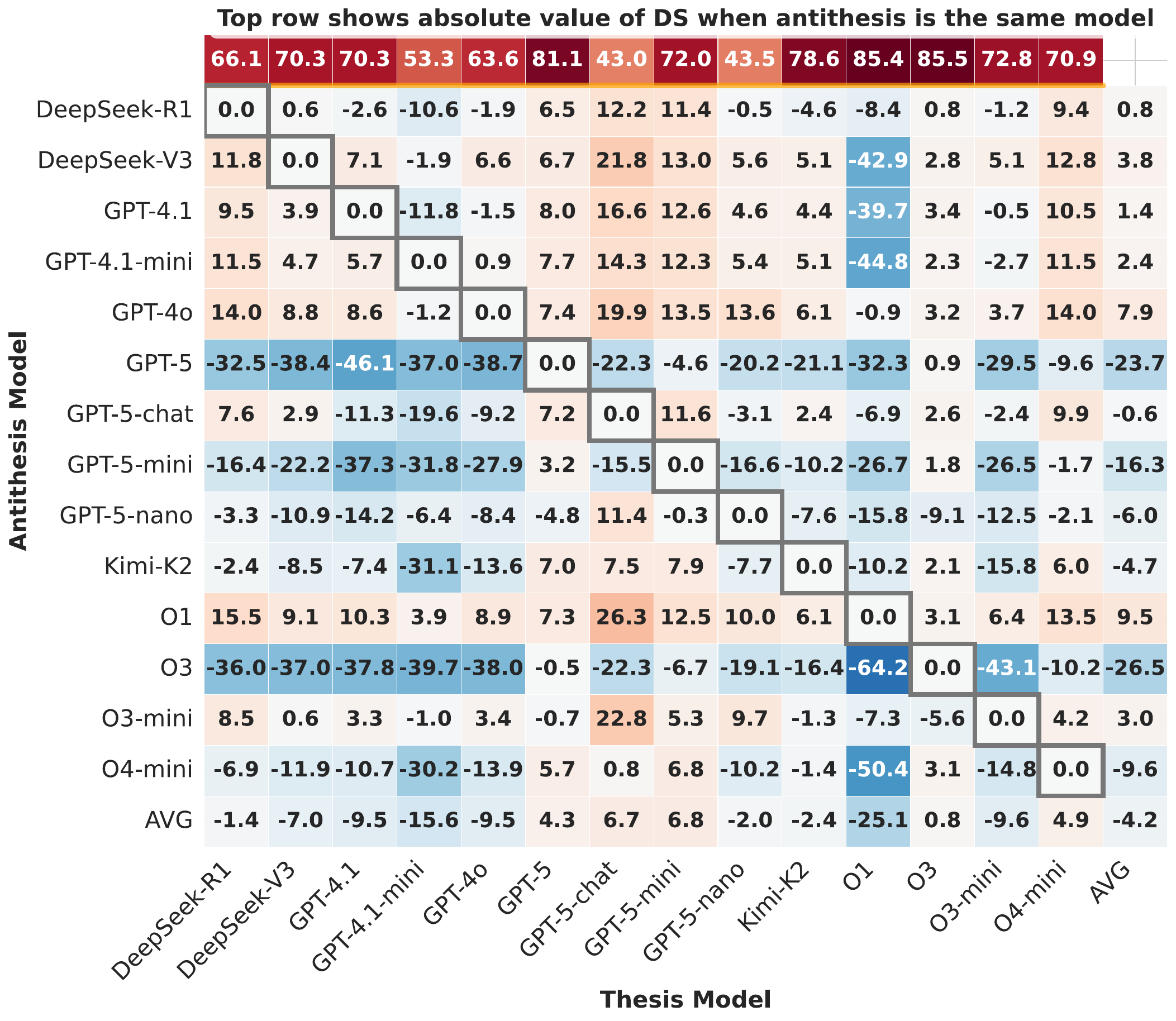}    
    \end{minipage}
    \captionsetup{font=normalsize}
    \caption{Heatmap graphs in cross-model dialectics for representative topics from MMLU benchmark showing OC (top left), $p_S$ (top right), and DS (bottom). Numbers in cells show $(\text{cross-X} - \text{self-X})$ values, where X is OC (top left graph), $p_S$ (top right graph), and DS (bottom graph). The top row in each graph shows the self-X values.}
    \label{fig:cross-model:mmlu}
\end{figure*}

The comparison between the performance of several models on the full MMLU benchmark and the five representative topics is presented in Table~\ref{table:app:mmlu:representative}. As shown, the differences are minimal, confirming that this subset serves as a reliable proxy for the full MMLU benchmark. Using these representative topics, we conducted cross-model dialectical evaluations, with the outcomes illustrated in Figure~\ref{fig:cross-model:mmlu}. While the absolute values differ from those observed for the GSM benchmark, the overall patterns and phenomena remain consistent with the trends discussed in Section~\ref{sec:eval:cross-model}.


\begin{table*}[htbp]
\centering
\caption{Thesis performance in MMLU benchmark and representative sub-topics}
\label{table:app:mmlu:representative}
\footnotesize
\begin{tabular}{lccc}
\toprule
Model & Overall MMLU Benchmark & Representative Topics & Difference \\
\midrule
O3             & 92.2 & 90.0 & -2.2 \\
GPT-5          & 92.2 & 90.1 & -2.1 \\
O1             & 91.1 & 89.8 & -1.3 \\
GPT-5-mini     & 89.3 & 86.9 & -2.4 \\
GPT-41         & 88.7 & 85.2 & -3.6 \\
O4-mini        & 88.6 & 86.5 & -2.1 \\
GPT-5-chat     & 88.2 & 84.3 & -4.0 \\
DeepSeek-V3    & 86.6 & 84.4 & -2.2 \\
GPT-4o         & 86.5 & 84.9 & -1.6 \\
GPT-5-nano     & 85.8 & 83.3 & -2.5 \\
O3-mini        & 85.0 & 83.4 & -1.6 \\
GPT-4.1        & 84.9 & 82.9 & -2.0 \\
O1-mini        & 81.8 & 81.6 & -0.2 \\
\bottomrule
\end{tabular}
\end{table*}

\section{MMLU Detailed \texorpdfstring{$\Delta$}{Delta} results}
\label{app:mmlu:full:drix}
\begin{figure*}[t]
    \centering
    \begin{minipage}[b]{0.9\textwidth}
        \centering
        \includegraphics[trim=0.25cm 0cm 0cm 0cm, clip,width=\textwidth]{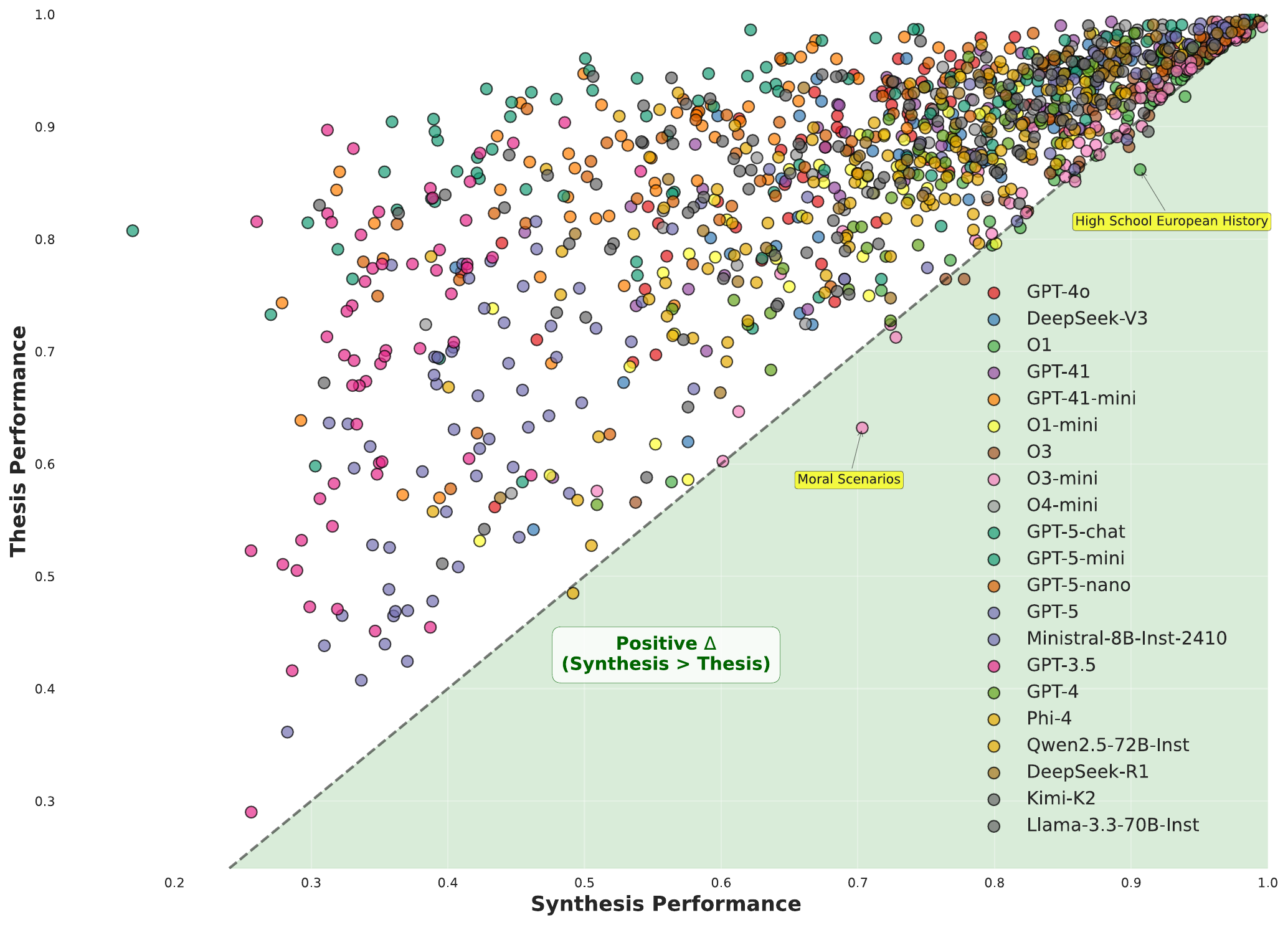}
    \end{minipage}   
    \caption{Detailed results of different models in different MMLU exams. Each circle shows the overall result of a model in a sub-topic in MMLU benchmark.}
    \label{fig:mmlu:full:delta}    
\end{figure*}

As discussed in Section~\ref{sec:eval:overall}, none of the models in our experiments consistently achieved a positive $\Delta$ score on average. In this section, we provide a more granular analysis of the MMLU results, detailing the performance of various models across the full spectrum of topics included in the MMLU benchmark. Figure~\ref{fig:mmlu:full:delta} illustrates these findings.
Interestingly, for certain topics, some models demonstrate improved reasoning performance and enter the green zone, where $\Delta > 0$. This suggests that under specific conditions, current models are capable of dialectical reasoning and can traverse the thesis-antithesis-synthesis structure to arrive at a higher-level conclusion. However, in the majority of cases, models fail to exhibit such reasoning capabilities and remain outside the green zone.
While the magnitude of $\Delta$ varies, this limitation underscores a lack of genuine reasoning—where reasoning is conceived as a dynamic process of confronting and integrating distinct and oppoings viewpoints. Instead, these models often appear to rely on pattern-matching behaviors to generate final answers, even in the form of a chain of tokens. In other words, although their initial correct responses may seem to reflect high reasoning capability, those responses can unravel when subjected to a more rigorous, dynamic dialectical procedure. That is why we need a more structured way to assess the reasoning capability of these models.

\section{Correlation Analysis of Dialectical Patterns for MMLU sub-exams}
\label{sec:app:correlation}

To capture the full spectrum of dependencies in dialectical reasoning performance, including non-linear and non-monotonic relationships, we employ distance correlation analysis~\cite{szekely2007measuring}. Distance correlation provides a dependency measure with the fundamental property that $\text{dCor}(X,Y) = 0$ if and only if $X$ and $Y$ are statistically independent. $\text{dCor}(X,Y)$ is computed through the following steps:
{
\[
\text{dCor}^2(X,Y) = \frac{\text{dCov}^2(X,Y)}{\sqrt{\text{dVar}(X)\,\text{dVar}(Y)}}, 
\]
\[
\quad
\text{where } \text{dCov}^2(X,Y) = \frac{1}{n^2} \sum_{k,l=1}^{n} A_{kl} B_{kl}.
\]
}
where $A_{kl}$ and $B_{kl}$ are double-centered Euclidean distance matrices~\cite{szekely2009brownian}.

\begin{figure*}[tb]
\centering
\includegraphics[width=\textwidth]{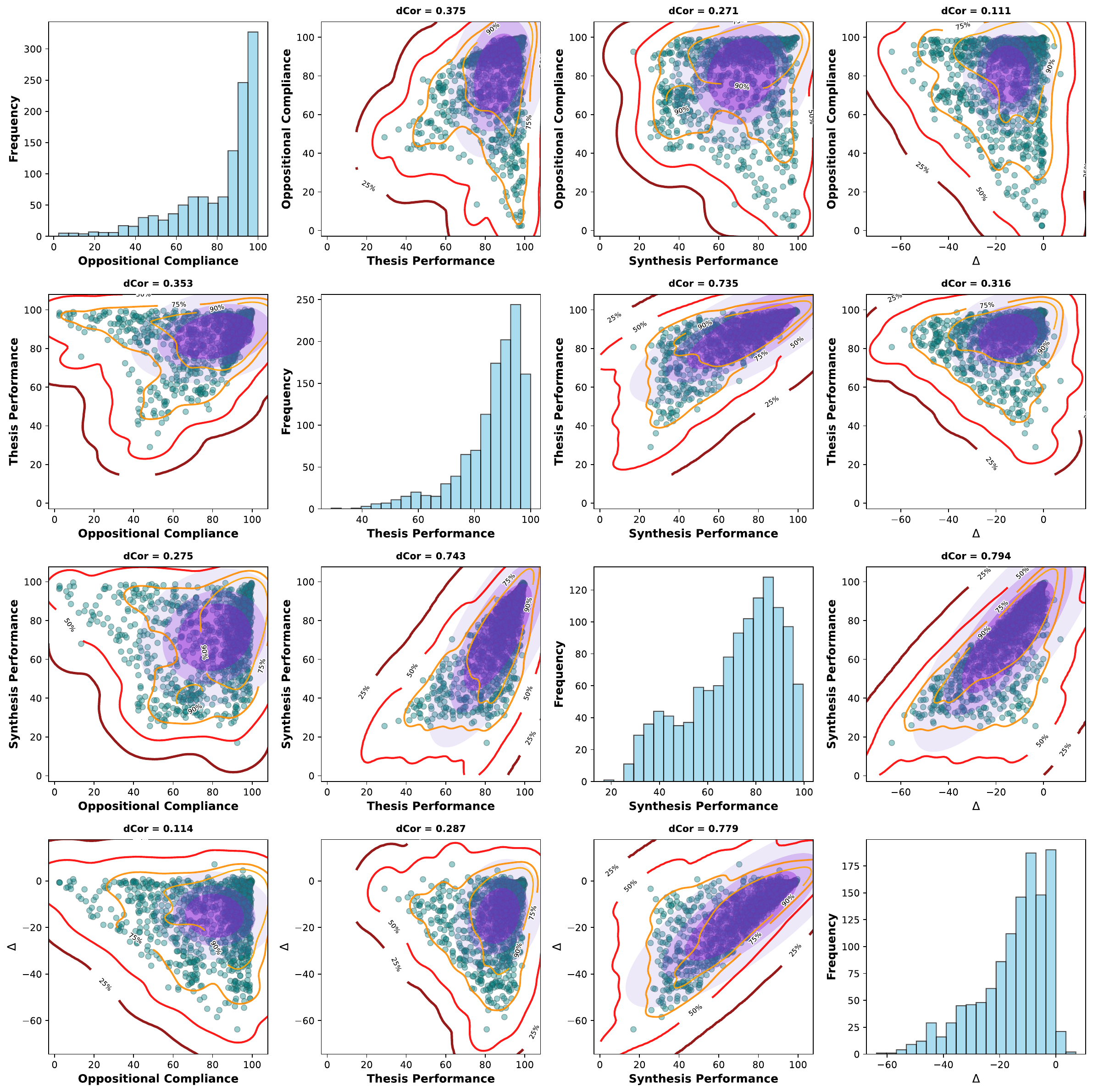}
\caption{
Distance Correlation Analysis of Dialectical Reasoning Performance for all tested LLMs. 
Distance correlation analysis across MMLU sub-topics reveals weak links between thesis performance and both $\Delta$ and OC, indicating that initial accuracy does not predict dialectical reasoning. As discussed in Section~\ref{sec:eval:overall}, low OC often leads to low $\Delta$, suggesting minimal deviation from the thesis when antitheses are weak. While thesis–synthesis correlation is stronger, it remains inconsistent—models with similar thesis scores can yield widely varying synthesis outcomes. 
Taken together with Section~\ref{sec:eval:overall}, these patterns suggest that high thesis accuracy, the conventional way to assess the reasoning capability of LLMs, does not by itself demonstrate genuine    reasoning. However, low thesis score reliably flags weak reasoning performance. 
}
\label{fig:distance_correlation:overall:4x4}
\end{figure*}

\begin{figure*}[tb]
\centering
\includegraphics[width=\textwidth]{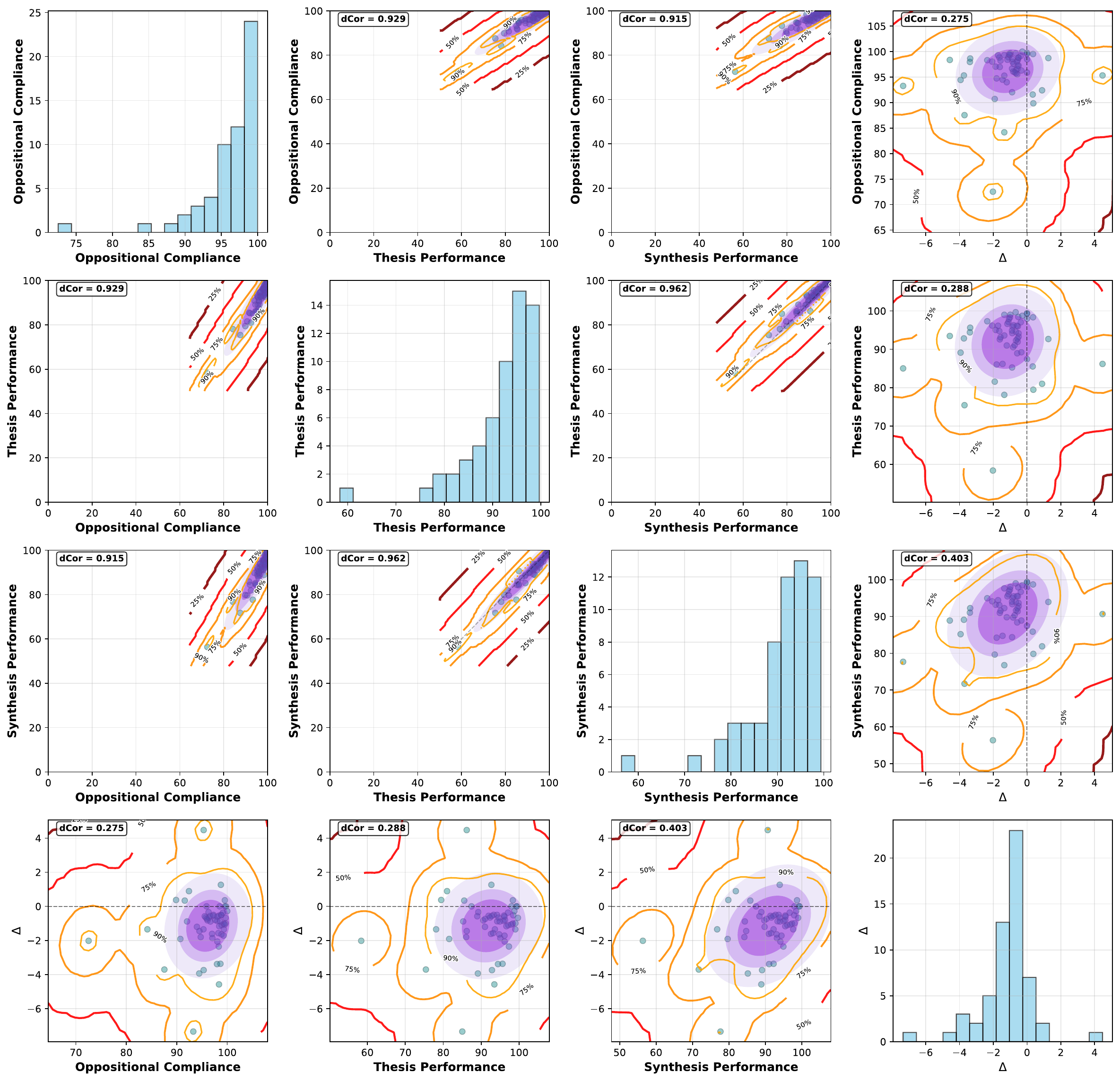}
\caption{
Distance Correlation Analysis of Dialectical Reasoning Performance for O1. The correlation values reveal notable patterns for the O1 model, a top performer. Thesis, OC, and synthesis scores exhibit strong, nearly linear correlations. However, the thesis–$\Delta$ subplot shows that in most cases, the model produces negative $\Delta$ values. This means that although O1 can generate antitheses that effectively challenge the thesis, it often fails to leverage these opposing views to reach a higher-order synthesis—indicating limited dialectical reasoning capability. Moreover, the narrow range of $\Delta$ values suggests that O1 tends to stay close to its original thesis, which explains the strong correlation between thesis and synthesis scores.
}
\label{fig:distance_correlation:o1}
\end{figure*}

\begin{figure*}[tb]
\centering
\includegraphics[width=\textwidth]{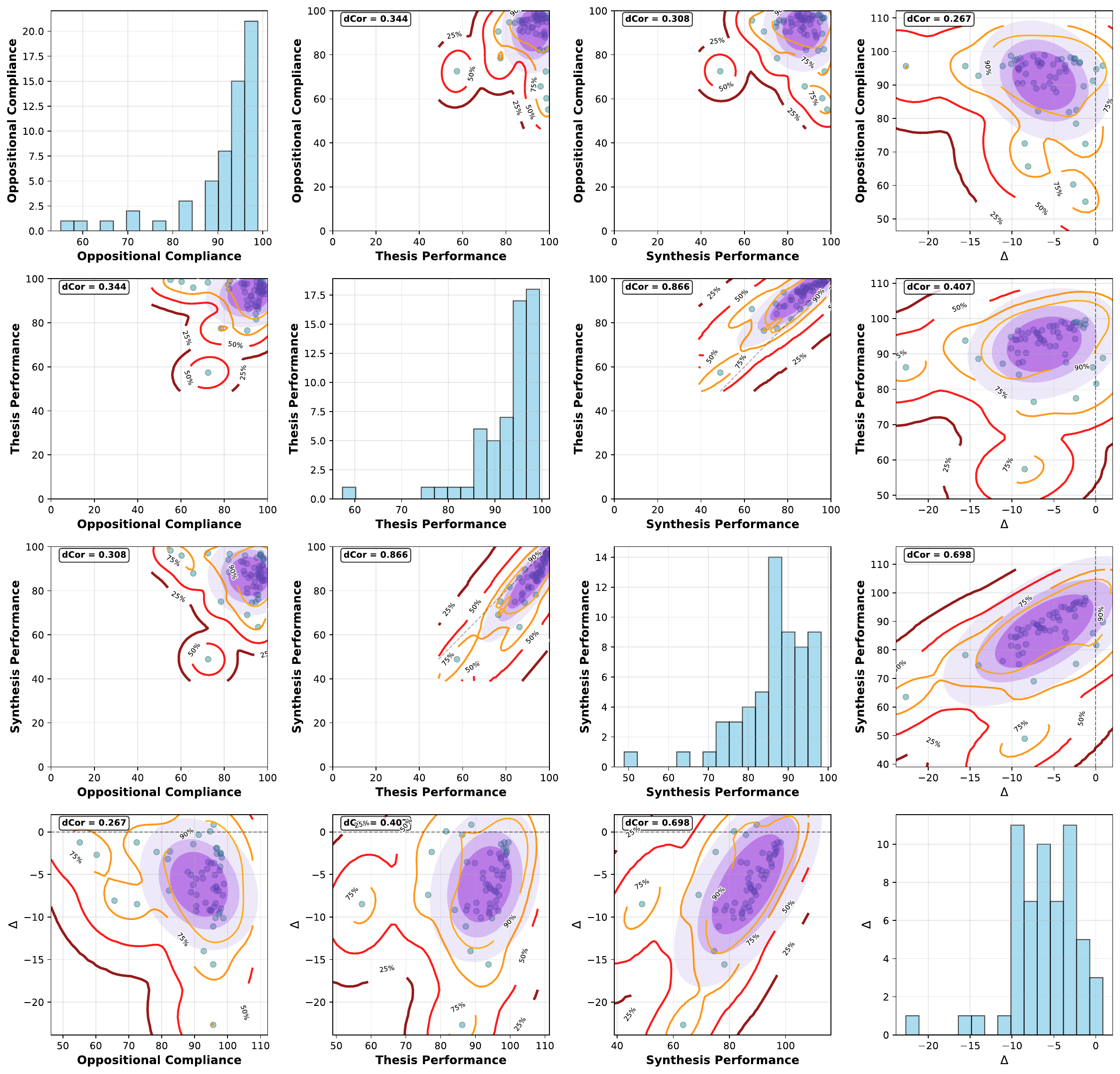}
\caption{
Distance Correlation Analysis of Dialectical Reasoning Performance for GPT-5.
The correlation matrix reveals several notable patterns in GPT-5’s dialectical reasoning behavior. While the model shows a strong correlation between thesis and synthesis performance, and a moderate link between synthesis and $\Delta$,its ability to perform dialectical reasoning and leveraging oppositional views remains limited.
The thesis–$\Delta$ correlation suggests that initial accuracy has some influence on dialectical improvement, but not reliably. More importantly, the weak correlations involving OC vs. $\Delta$ and OC vs. synthesis indicate that GPT-5 struggles to transform strong antitheses into meaningful synthesis. In other words, while the model can generate counterpoints, it often fails to use them constructively to reach higher-order reasoning.
This behavior reflects a pattern where GPT-5’s synthesis tends to align closely with its thesis, rather than evolving through dialectical engagement. The model’s reasoning trajectory appears more static than dynamic, reinforcing the need to evaluate models not just by outcome accuracy, but by their ability to reason through structured and dynamic process.
}
\label{fig:distance_correlation:gpt5}
\end{figure*}

\begin{figure*}[tb]
\centering
\includegraphics[width=\textwidth]{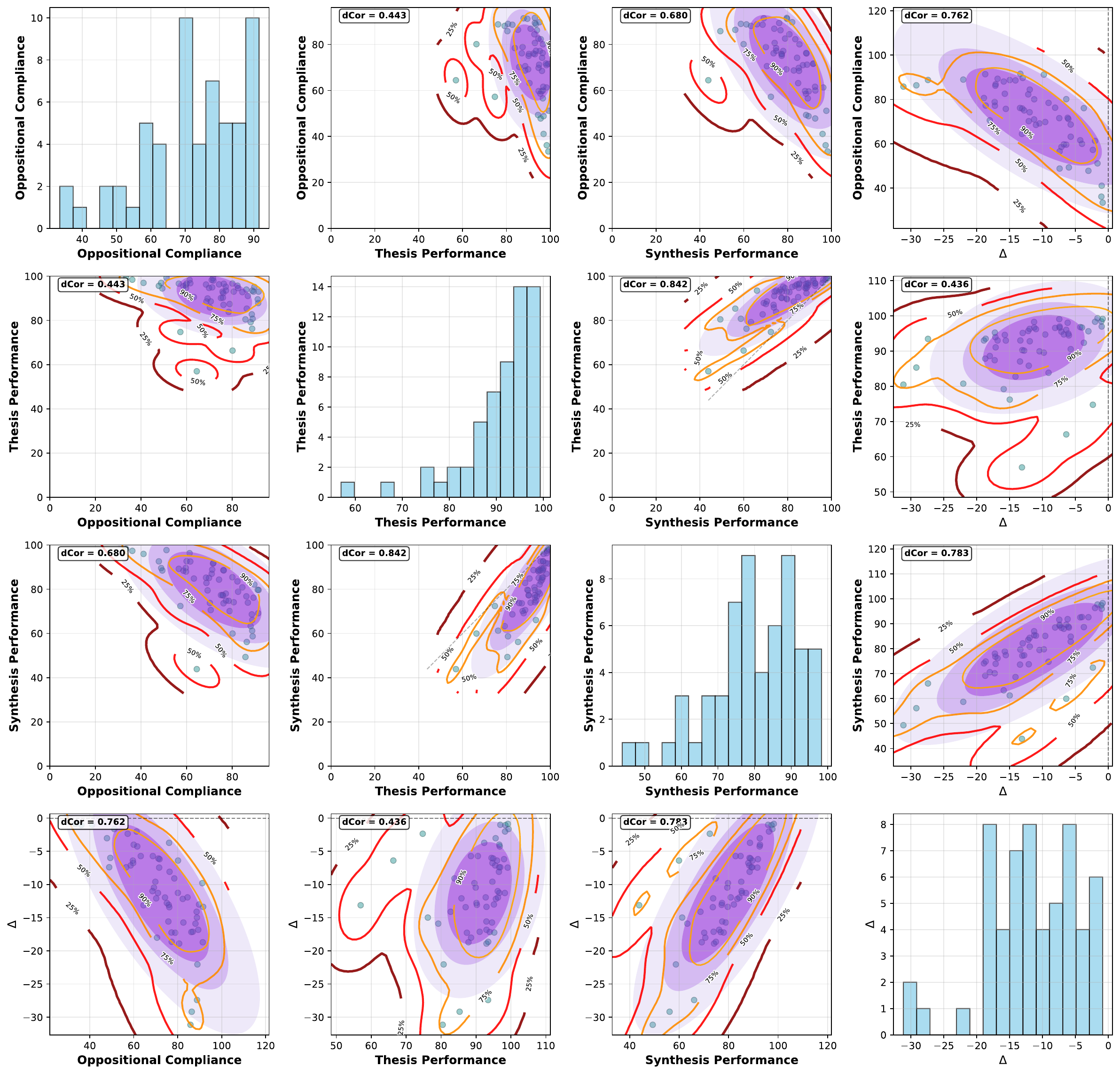}
\caption{
Distance Correlation Analysis of Dialectical Reasoning Performance for DeepSeek R1.
The correlation matrix for DeepSeek R1 reveals a more moderate dialectical reasoning profile compared to top-performing models like O1 and GPT-5. In our broader evaluation (Section~\ref{sec:eval:overall}), DeepSeek R1 ranks as a middle-tier model, with both its synthesis and OC scores noticeably lower than those of O1 and GPT-5. This is also reflected in the histogram plots: while the thesis scores peak near the maximum, the synthesis scores shift downward, clustering around the 80-point mark.
Although DeepSeek R1 shows moderate correlations between thesis, synthesis, and $\Delta$, its ability to make use of opposing views is limited. Similar to the general trend, the weak correlation between thesis and OC suggests that the presence of meaningful opposition in antitheses is not strongly related to the quality of the initial thesis. More importantly, the model often fails to integrate these opposing views into a higher-level synthesis. In many cases, the synthesis either closely mirrors the thesis or drops in quality—sometimes by as much as 30/100 points—highlighting the model’s limited capacity for dynamic, dialectical reasoning.
Overall, DeepSeek R1’s performance reinforces the importance of evaluating models not just by their initial accuracy, but by how effectively they engage with and evolve through structured reasoning processes. Its behavior underscores the need for metrics that go beyond static correctness to assess genuine reasoning capability.
}
\label{fig:distance_correlation:r1}
\end{figure*}

\begin{figure*}[tb]
\centering
\includegraphics[width=\textwidth]{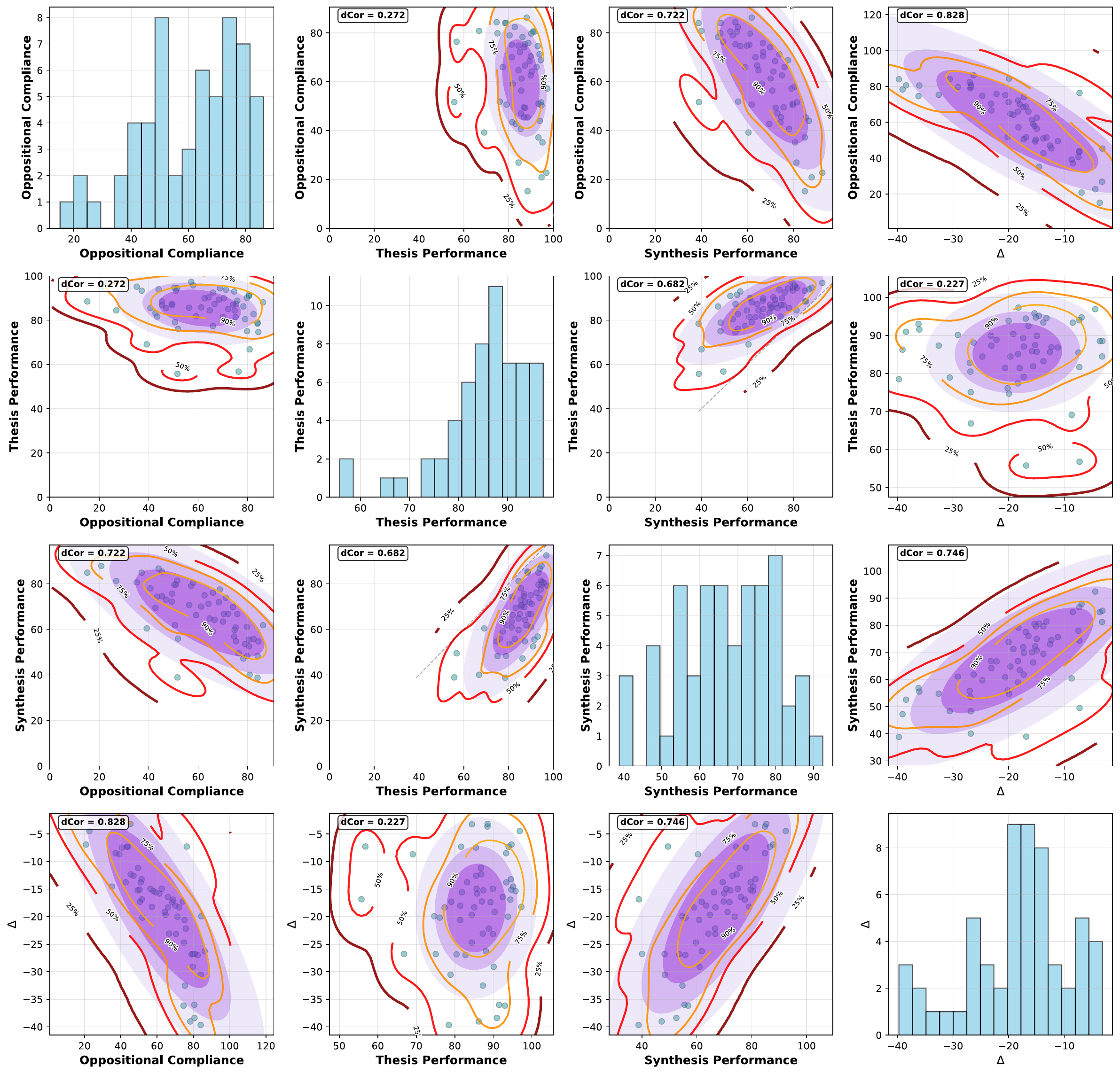}
\caption{
Distance Correlation Analysis of Dialectical Reasoning Performance for Qwen2.5-70B-Instruct.
The correlation matrix for Qwen2.5-70B-Instruct reveals a notably weaker dialectical reasoning profile, placing it below DeepSeek R1 in our broader evaluation. Both its synthesis and Oppositional Compliance (OC) scores are lower, as confirmed by the histogram plots: synthesis scores are widely dispersed and skewed toward lower values, while OC—measuring how often antitheses oppose the thesis—is also low and unevenly distributed.
The overall correlation pattern resembles that of a mid-performing model like R1, but with more extreme values. For instance, the model’s reasoning performance can degrade by as much as 40 points (out of 100), and its OC scores can drop below 20—indicating failure to generate meaningful opposition.
Overall, Qwen2.5-70B-Instruct demonstrates how even large-scale models struggle with reasoning when perceived through a dynamic dialectical setting, reinforcing the need for evaluation frameworks that go beyond static accuracy.
}
\label{fig:distance_correlation:qwen2.5-72b-inst}
\end{figure*}

\begin{figure*}[tb]
\centering
\includegraphics[width=\textwidth]{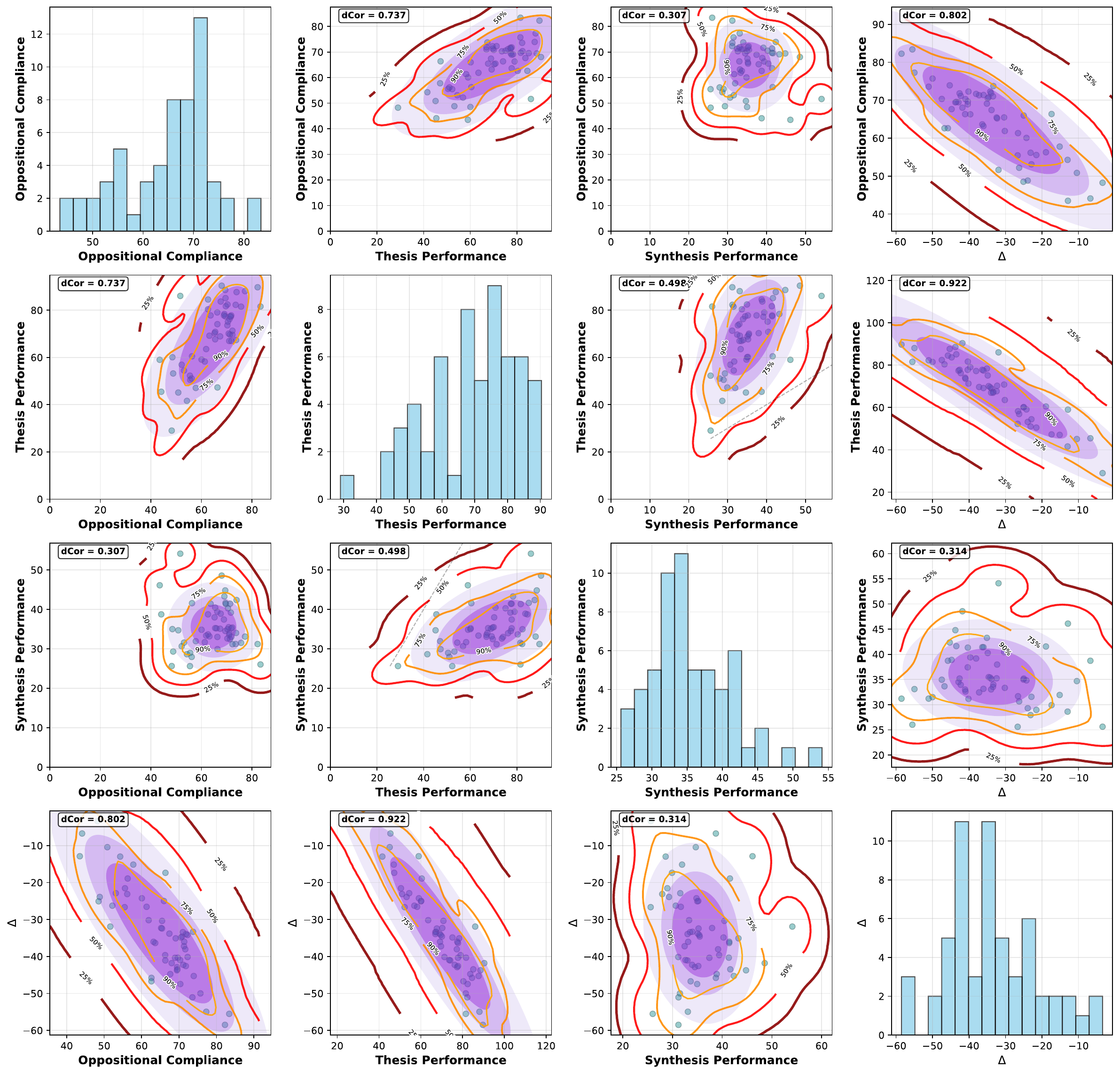}
\caption{
Distance Correlation Analysis of Dialectical Reasoning Performance for GPT-3.5.
The correlation matrix for GPT-3.5 reveals the weakest dialectical reasoning profiles among all evaluated models. Across the board, its synthesis and $\Delta$ scores are low, and the histogram plots confirm this: synthesis scores are broadly scattered and skewed toward the lower end. One immediate different pattern comapred to other models is the strong correlation between thesis and $\Delta$. The higher the thesis performance is the lower the $\Delta$ will be. The thesis-OC graph can shed light on this. With increased thesis performance, GPT-3.5 shows increased OC, and consequently with a model that lacks deep reasoning capabilities, it ends up to a lower quality synthesis. As graphs show, the model's performance can degrade as much as 60/100 points in some scenarios. Overall, GPT-3.5 demonstrates limited capacity for dynamic reasoning. 
}
\label{fig:distance_correlation:gpt3.5}
\end{figure*}

\begin{figure*}[tb]
\centering
\includegraphics[width=\textwidth]{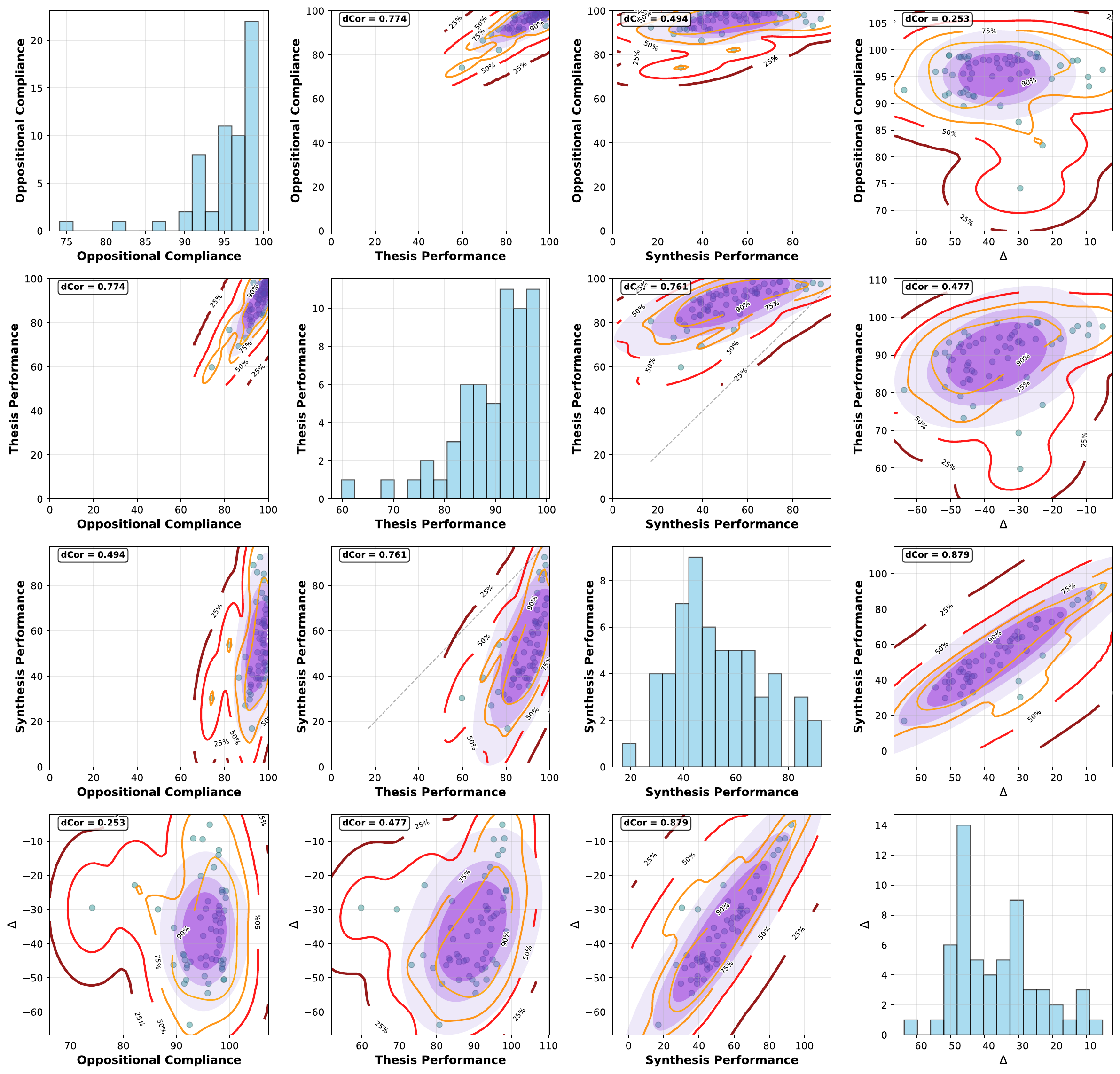}
\caption{
Distance Correlation Analysis of Dialectical Reasoning Performance for GPT-5-chat.
The correlation matrix for GPT-5-chat reveals a weak dialectical reasoning profile, placing it among the lowest-performing models in our broader evaluation. Both synthesis and $\Delta$ scores are low, as confirmed by the histogram plots: synthesis scores are broadly scattered and skewed toward the lower range, while $\Delta$ values show limited improvement and frequent degradation.
When opposition is present, the model often fails to integrate it into a higher-order synthesis, resulting in reasoning trajectories that either stagnate or degrade (more than 60/100 points degradations in some cases).
A notable contrast emerges when comparing GPT-5-chat to GPT-3.5, another low-performing model. GPT-3.5 exhibits a strong correlation between thesis and $\Delta$, where higher thesis scores often lead to greater synthesis degradation due to increased OC. In contrast, GPT-5-chat does not show a strong thesis–$\Delta$ correlation, yet it does exhibit a high correlation between thesis and OC. This suggests that while GPT-5-chat generates more opposition as thesis performance increases, its reasoning trajectory falls apart ending up to have very low synthesis scores. 
Overall, GPT-5-chat demonstrates limited capacity for dynamic reasoning, yet again, reinforcing the need for evaluation frameworks that go beyond static accuracy and assess how models engage with and evolve through structured dynamic settings.
}
\label{fig:distance_correlation:gpt-5-chat}
\end{figure*}

In what follows we present the distance correlation analysis in MMLU benchmark and its sub-exams. IN particular, we present resutls for (1) overall LLMs: Figure~\ref{fig:distance_correlation:overall:4x4}, (2) two of top performing models, O1 and GPT-5 (Figure~\ref{fig:distance_correlation:o1} and Figure~\ref{fig:distance_correlation:gpt5}, respectively), (3) two of the middle performing ones, DeepSeek-R1 and Qwen2.5-72B-Instruct (Figure~\ref{fig:distance_correlation:r1} and Figure~\ref{fig:distance_correlation:qwen2.5-72b-inst}, respectively) (4) two of the low performing ones, GPT-3.5, GPT-5-chat (Figure~\ref{fig:distance_correlation:gpt3.5} and Figure~\ref{fig:distance_correlation:gpt-5-chat}, respectively).

\soheilhead{General Notes about the Figures}
\begin{itemize}[leftmargin=0.4cm]
    \item Matrix Structure: Figures show a symmetric correlation matrix displaying all pairwise relationships between four key variables: OC (Oppositional Compliance), Thesis Performance, Synthesis Performance, and $\Delta$ (dialectical gap). 
    \item Diagonal Elements: Histograms show the distribution of each variable, with bins displaying frequency patterns. 
    \item  Contour Visualization: Two overlays per scatter plot: (1) Purple/pink confidence ellipses at 50\%, 75\%, and 95\% dependence levels, with darker shades indicating stronger confidence regions; (2) Orange/red density contour lines labeled 25\%, 50\%, 75\%, 90\% representing data concentration percentiles, from dark red (highest density core) to light orange (outer data boundary). 
    \item Distance Correlation Values: Each scatter plot displays dCor values (0-1 scale) in the title, capturing both linear and non-linear dependencies. Contour shapes reveal dependency patterns beyond simple linear correlations. 
    \item Symmetry: The matrix exploits correlation symmetry—relationships below the diagonal mirror those above, providing pairwise analysis.
\end{itemize}

\end{document}